\documentclass{article}

\usepackage{template_v2}

\usepackage[utf8]{inputenc} %
\usepackage[T1]{fontenc}    %
\usepackage{hyperref}       %
\usepackage{url}            %
\usepackage{booktabs}       %
\usepackage{amsfonts}       %
\usepackage{nicefrac}       %
\usepackage{microtype}      %
\usepackage{xcolor}         %
\usepackage{graphicx}
\usepackage{xspace}         %
\usepackage{booktabs}
\usepackage{amsmath}
\usepackage{wrapfig}
\usepackage[breakable]{tcolorbox}
\tcbuselibrary{skins, breakable}
\usepackage{newunicodechar}
\usepackage[frozencache,cachedir=_minted]{minted}
\setminted{
  fontsize=\normalsize,
  breaklines=true,
  breakanywhere=true,
  breaksymbolleft={},
  breakautoindent=false,
  tabsize=2,
}

\newunicodechar{ℝ}{\ensuremath{\mathbb{R}}}
\newunicodechar{ℕ}{\ensuremath{\mathbb{N}}}
\newunicodechar{∃}{\ensuremath{\exists}}
\newunicodechar{∀}{\ensuremath{\forall}}
\newunicodechar{∧}{\ensuremath{\wedge}}
\newunicodechar{₁}{\ensuremath{_1}}
\newunicodechar{₂}{\ensuremath{_2}}
\newunicodechar{₃}{\ensuremath{_3}}
\newunicodechar{₄}{\ensuremath{_4}}
\newunicodechar{₅}{\ensuremath{_5}}
\newunicodechar{⊢}{\ensuremath{\vdash}}

\usepackage{fvextra}
\DeclareUnicodeCharacter{211D}{R}        %
\DeclareUnicodeCharacter{2115}{N}        %
\DeclareUnicodeCharacter{2102}{C}        %
\DeclareUnicodeCharacter{2200}{forall}   %
\DeclareUnicodeCharacter{2203}{exists}   %
\DeclareUnicodeCharacter{2227}{and}      %
\DeclareUnicodeCharacter{2228}{or}       %
\DeclareUnicodeCharacter{27E8}{<}        %
\DeclareUnicodeCharacter{27E9}{>}        %
\DeclareUnicodeCharacter{00AC}{not}      %
\DeclareUnicodeCharacter{2192}{->}       %
\DeclareUnicodeCharacter{2264}{<=}       %
\DeclareUnicodeCharacter{2265}{>=}       %
\DeclareUnicodeCharacter{22A2}{|-}       %
\DeclareUnicodeCharacter{2081}{_1}       %
\DeclareUnicodeCharacter{2082}{_2}       %
\DeclareUnicodeCharacter{2083}{_3}       %
\DeclareUnicodeCharacter{2084}{_4}       %
\DeclareUnicodeCharacter{2085}{_5}       %
\DeclareUnicodeCharacter{00B2}{^2}       %
\DeclareUnicodeCharacter{2081}{1}   %
\DeclareUnicodeCharacter{2082}{2}   %
\DeclareUnicodeCharacter{2083}{3}   %
\DeclareUnicodeCharacter{2084}{4}   %
\DeclareUnicodeCharacter{2085}{5}   %
\DeclareUnicodeCharacter{00B2}{2}   %
\DeclareUnicodeCharacter{2124}{\ensuremath{\mathbb{Z}}}   %
\DeclareUnicodeCharacter{211A}{\ensuremath{\mathbb{Q}}}   %
\DeclareUnicodeCharacter{2102}{\ensuremath{\mathbb{C}}}   %
\DeclareUnicodeCharacter{27E8}{\ensuremath{\langle}}      %
\DeclareUnicodeCharacter{27E9}{\ensuremath{\rangle}}      %
\DeclareUnicodeCharacter{2218}{\ensuremath{\circ}}        %
\DeclareUnicodeCharacter{2264}{\ensuremath{\leq}}         %
\DeclareUnicodeCharacter{2260}{\ensuremath{\neq}}         %
\DeclareUnicodeCharacter{2124}{\ensuremath{\mathbb{Z}}}   %
\DeclareUnicodeCharacter{22A2}{\ensuremath{\vdash}}      %
\DeclareUnicodeCharacter{2080}{\ensuremath{_0}}          %
\DeclareUnicodeCharacter{2081}{\ensuremath{_1}}          %
\DeclareUnicodeCharacter{2082}{\ensuremath{_2}}          %
\DeclareUnicodeCharacter{2083}{\ensuremath{_3}}          %
\DeclareUnicodeCharacter{2084}{\ensuremath{_4}}          %
\DeclareUnicodeCharacter{2085}{\ensuremath{_5}}          %
\DeclareUnicodeCharacter{2086}{\ensuremath{_6}}          %
\DeclareUnicodeCharacter{2087}{\ensuremath{_7}}          %
\DeclareUnicodeCharacter{2088}{\ensuremath{_8}}          %
\DeclareUnicodeCharacter{2089}{\ensuremath{_9}}          %
\DeclareUnicodeCharacter{2194}{\ensuremath{\leftrightarrow}}   %
\DeclareUnicodeCharacter{27E8}{\ensuremath{\langle}}           %
\DeclareUnicodeCharacter{27E9}{\ensuremath{\rangle}}           %
\DeclareUnicodeCharacter{00B7}{\ensuremath{\cdot}}             %
\DeclareUnicodeCharacter{00AC}{\ensuremath{\neg}}              %
\DeclareUnicodeCharacter{2192}{\ensuremath{\to}}               %
\DeclareUnicodeCharacter{2190}{\ensuremath{\leftarrow}}        %
\DeclareUnicodeCharacter{2218}{\ensuremath{\circ}}             %
\DeclareUnicodeCharacter{2260}{\ensuremath{\neq}}              %
\DeclareUnicodeCharacter{2264}{\ensuremath{\leq}}

\usepackage{array}
\usepackage{xcolor}

\usepackage[table]{xcolor}

\usepackage{arydshln}

\usepackage{multirow}
\usepackage{natbib}

\usepackage{pifont}

\definecolor{plotblue}{HTML}{567CBA}
\definecolor{plotorange}{HTML}{E79A64}
\definecolor{plotgreen}{HTML}{55A868}
\definecolor{plotred}{HTML}{CE585C}
\definecolor{plotpurple}{HTML}{8172B3}
\definecolor{plotpink}{HTML}{DA8BC3}
\definecolor{plotmustard}{HTML}{CCB974}
\definecolor{plotcyan}{HTML}{64B5CD}

\definecolor{acl-blue}{RGB}{52, 109, 187}
\definecolor{softgreen}{HTML}{2E8B57}

\hypersetup{
  colorlinks=true,
  linkcolor=softgreen, %
  citecolor=acl-blue,
  urlcolor=acl-blue
}

\usepackage{subcaption}

\newcommand{\method}{Pythagoras-Prover\xspace}

\title{%
  \hspace*{3.2em}%
  \makebox[0pt][r]{%
    \smash{\raisebox{\dimexpr-0.5\height-0.2\baselineskip\relax}{%
      \includegraphics[height=2.9\baselineskip]{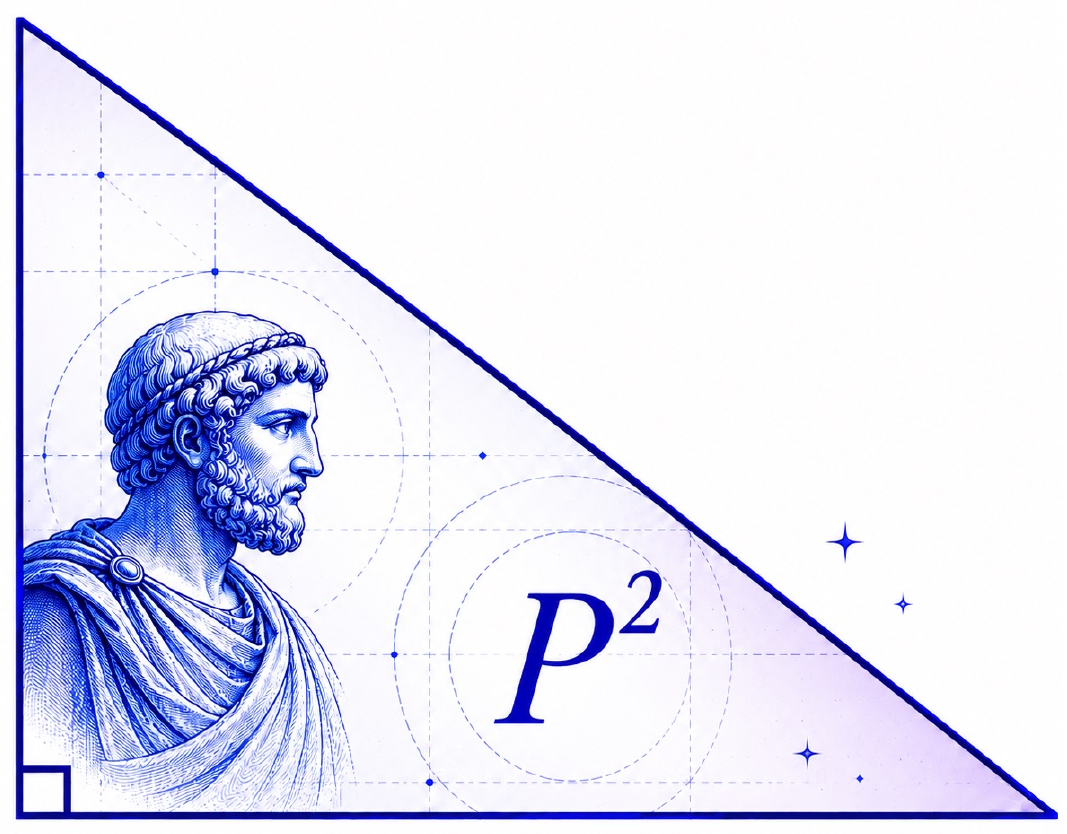}%
    }}%
    \hspace{-1.0em}%
  }%
  Pythagoras-Prover: Advancing Efficient Formal \\
  \hspace*{2em} Proving via Augmented Lean Formalisation%
}

\renewcommand{\shorttitle}{\emph{\method}}

\date{}

\author{%
  \textbf{Joshua Ong Jun Leang}\thanks{Corresponding author: \texttt{j.ong25@imperial.ac.uk}}$^{\,p^2,a^2}$ \quad
  \textbf{Zheng Zhao}$^{a^2}$ \quad
  \textbf{Mihaela C\u{a}t\u{a}lina Stoian}$^{p^2}$ \quad
  \textbf{Qiyuan Xu}$^{b^2}$ \quad \\
  \textbf{Haonan Li}$^{c^2}$ \quad
  \textbf{Wenda Li}$^{a^2}$ \quad
  \textbf{Shay B. Cohen}$^{a^2}$ \quad
  \textbf{Eleonora Giunchiglia}$^{p^2}$ \\[4pt]
  $^{p^2}$Imperial College London \quad
  $^{a^2}$University of Edinburgh \quad
  $^{b^2}$Nanyang Technological University \quad
  $^{c^2}$MBZUAI
}

\begin{document}

\maketitle

\begin{abstract}

Modern Lean theorem provers achieve strong performance only with substantial training and inference compute. This cost is driven in part by the scarcity of verified proof data and by the long reasoning traces required for formal proof search, which make both supervised fine-tuning and sampling expensive. We introduce \method, a compute-efficient open-source family of Lean theorem provers designed to deliver strong performance under practical compute budgets. The family spans two generation paradigms: two autoregressive models with 4B and 32B parameters, and a first proof-of-concept diffusion-based theorem-proving model (4B), which iteratively refines Lean proofs at inference time. 
To make training more efficient, we construct a Lean-verified corpus stratified into easy, medium, and hard problems and use it for curriculum supervised fine-tuning, allowing the models to acquire proof skills progressively from shorter and simpler proofs to longer and more difficult ones. During supervised fine-tuning, we further apply a dynamic proof-reasoning filtering scheme that preserves informative proof traces while ensuring each training instance fits within an 8k-token context budget. We further introduce \emph{Augmented Lean Formalisation} (ALF), which expands scarce verified corpora into  variants of formal statements; these variants are then populated through self-distillation, providing additional training signal without requiring every mutated instance to be formally verified. By perturbing known problems while preserving their formal character, ALF exposes the model to structured variants of verified problems, reducing reliance on any single statement's surface form.
Empirically, \method demonstrates strong performance across model scales. Most notably, \method-4B surpasses DeepSeek-Prover-V2-671B at pass@32 on MiniF2F-Test ($82.4\% \to 86.1\%$), despite using roughly $167\times$ fewer parameters. Scaling to 32B further yields state-of-the-art performance among open-source neural theorem provers, with \method-32B attaining $93.0\%$ on MiniF2F-Test and solving $93$ of $672$ problems on PutnamBench. We additionally release \emph{MiniF2F-ALF}, an ALF-mutated     contamination-sensitive perturbation benchmark on which every evaluated model loses accuracy; on this split, \method-32B remains the strongest evaluated prover, while \method-4B reaches parity with Goedel-Prover-V2-32B, the prior state of the art. Together, these results show that strong Lean theorem proving need not rely exclusively on frontier-scale models. Our models, data, and benchmark are released at~\href{https://huggingface.co/Pythagoras-LM}{https://huggingface.co/Pythagoras-LM}.

\begin{figure}[ht]
    \centering
    \includegraphics[width=\columnwidth]{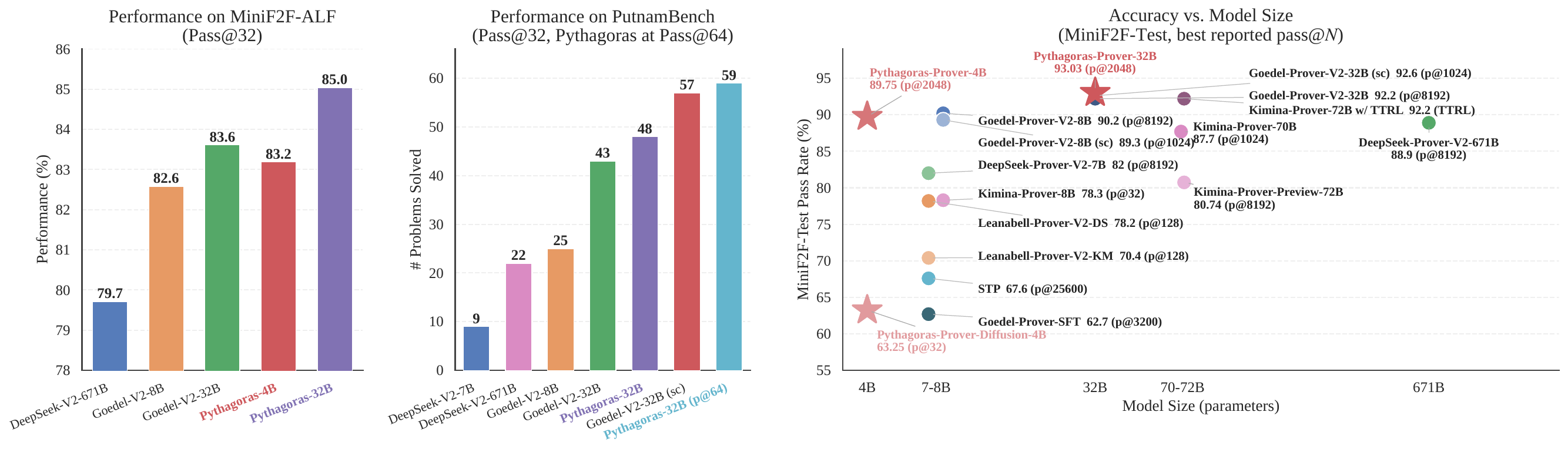}
    \caption{Performance comparison across proving tasks. The left and middle panels show results under a limited Pass@32 inference budget on MiniF2F-ALF and PutnamBench, respectively; for PutnamBench we additionally include Pythagoras-Prover at Pass@64, which incurs an inference cost comparable to Goedel-Prover-V2 with self-correction (see \S\ref{sec:results} for detailed explanation). The right panel plots best-reported MiniF2F-Test pass rate against model size (log scale), with the inference budget for each result shown in parentheses; stars mark our models (\method), and ``(sc)'' denotes self-correction.}
    \label{fig:prover_fig}
\end{figure}

\end{abstract}

\section{Introduction}

Recent progress in mathematical reasoning has made large language models (LLMs) considerably more capable, yet not reliably correct. Although modern LLMs can solve complex problems end-to-end, their reasoning remains susceptible to hallucinations and subtle logical errors that natural-language inspection cannot reliably detect~\citep{leang-etal-2025-comat, lyu-etal-2023-faithful}. Automated Theorem Proving (ATP) addresses this limitation by grounding model outputs in interactive proof assistants such as Lean~\citep{DBLP:conf/cade/MouraKADR15}, Isabelle~\citep{DBLP:books/sp/Paulson94} and Rocq~\citep{CoquandHuet1988CoC}, whose deterministic type-checkers reject any argument that is not mechanically verifiable. This formal grounding eliminates an entire class of hallucinations and yields reasoning whose correctness can be audited step by step.

In recent years, frontier systems such as DeepMind's AlphaProof and AlphaGeometry~\citep{DBLP:journals/jmlr/ChervonyiTOYNMJ25, 1062014} and ByteDance's Seed-Prover~\citep{chen2025seed} have demonstrated that AI systems can attain International Mathematical Olympiad (IMO) medal-level performance. In the open-source community, DeepSeek-Prover-V2~\citep{DBLP:journals/corr/abs-2504-21801}, Kimina-Prover~\citep{wang2025kiminaproverpreviewlargeformal}, and Goedel-Prover-V2~\citep{lin2025goedelproverv2scalingformaltheorem} have reported strong results on standard benchmarks including MiniF2F~\citep{DBLP:journals/corr/abs-2109-00110} and PutnamBench~\citep{tsoukalas2024putnambenchevaluatingneuraltheoremprovers}. These successes, however, are typically achieved either through very large models with hundreds of billions of parameters, or through computationally intensive inference that relies on elaborate search procedures, self-correction, or large sampling budgets. State-of-the-art ATP therefore remains largely inaccessible to researchers and practitioners without substantial compute, and a sizeable gap persists between small open-source provers and their largest counterparts.

In this work, we release \method, a compute-efficient family of open-source theorem provers for Lean~4 that challenges the assumption that stronger formal reasoning requires frontier-scale models. At 4B parameters, \method-4B surpasses DeepSeek-Prover-V2-671B~\citep{DBLP:journals/corr/abs-2504-21801} at pass@32 on MiniF2F-Test ($82.4\% \to 86.1\%$), despite being roughly $167\times$ smaller. At 32B parameters, \method-32B achieves the strongest reported performance among open-weight neural Lean provers evaluated without self-correction, obtaining the highest pass rate on MiniF2F-Test while relying on standard restart sampling rather than inference-time corrective procedures. Beyond autoregressive proving, the family includes \method-Diffusion, to our knowledge the first proof-of-concept diffusion-based theorem-proving model for Lean. Under a matched 4B setting on the same self-distillation corpus and evaluation harness, \method-Diffusion trails \method-4B in raw pass@32 accuracy but generates proofs $2.58\times$ faster on the same hardware, placing it on the throughput side of an emerging accuracy-efficiency frontier for formal theorem proving. %

Underlying all our releases is a compute-frugal data recipe for Lean theorem proving. We first build a Lean-verified seed corpus of roughly 800K instances using predominantly open models with 30B parameters or fewer, with a single 235B model as the only larger component. This verified corpus is stratified into easy, medium, and hard difficulty tiers, which enables curriculum supervised fine-tuning from shorter and simpler proofs to longer and more difficult ones. We then introduce \emph{Augmented Lean Formalisation} (ALF), which takes each verified instance and produces structured formal variants across five mutation types: simplification, generalisation, lemma proposal, proof-step decomposition, and reformulation. Because ALF uses a lightweight sanity check rather than Lean-verifying every mutated instance, it decouples corpus expansion from verifier throughput and allows for the expansion of the seed corpus by roughly $2.5\times$. These ALF variants are then used in two ways: first, as self-distillation inputs that enrich post-training for both the autoregressive and diffusion provers; and second, as the basis for \emph{MiniF2F-ALF}, where the same mutation operator is applied to MiniF2F-Test to probe whether performance transfers across nearby formal variants. Contemporary provers degrade substantially on this benchmark, suggesting that strong performance on the original split does not always transfer to structured formal variants. Thus, the 4B and 32B autoregressive provers, the diffusion prover, and the companion benchmark all follow from the same pipeline: verified seed construction enables curriculum learning, ALF expands the seed into structured variants, self-distillation turns those variants into training signal, and the same perturbation mechanism yields a contamination-sensitive perturbation benchmark.

In summary, our contributions are as follows:

\begin{itemize}
\item \textbf{A compute-efficient family of Lean theorem provers.}
We release \method, a family of open-source Lean~4 theorem provers spanning two generation paradigms: autoregressive provers at 4B and 32B parameters, and \method-Diffusion-4B, to our knowledge the first proof-of-concept diffusion-based theorem-proving model. \method-4B surpasses DeepSeek-Prover-V2-671B on MiniF2F-Test at pass@32 ($82.4\% \to 86.1\%$), despite being roughly $167\times$ smaller, while \method-32B achieves the strongest reported performance among open-weight neural Lean provers evaluated without self-correction.
\item \textbf{A compute-frugal Lean data pipeline.}
We construct a Lean-verified corpus of roughly 800K instances stratified into easy, medium, and hard tiers. We then introduce \emph{Augmented Lean Formalisation} (ALF), a structured mutation scheme that expands this corpus by roughly $2.5\times$ using lightweight sanity checks rather than per-instance Lean verification. Applying the same mutation operator to MiniF2F-Test yields \emph{MiniF2F-ALF}, a companion benchmark for probing transfer across nearby formal variants of benchmark statements.

\item \textbf{An efficient adaptation recipe for Lean-proving.}
We combine parameter-efficient supervised fine-tuning, curriculum learning, self-distillation, and reinforcement learning from Lean verification. During supervised fine-tuning, a dynamic proof--reasoning filter preserves informative reasoning traces while keeping each instance within an 8K-token context budget. The same post-training corpus trains both the autoregressive and diffusion provers, allowing a controlled comparison between generation paradigms.
\end{itemize}

\section{Methodology}

In this section we present the training pipeline of \method in detail. We first describe our framework for generating the Lean-verified \emph{seed} corpus (\S\ref{sec:synthetic_data}). We then present the training algorithm that turns this corpus into our base autoregressive provers (\S\ref{sec:training_algorithm}). Finally, we introduce \emph{Augmented Lean Formalisation} (ALF), a structured mutation operator that scales the corpus beyond what per-instance Lean verification can throughput, and the self-distillation stage built on it (\S\ref{sec:alf}); the same self-distillation corpus is then used to train a diffusion-based prover (\S\ref{sec:scaling_diffusion}).

\begin{figure}[t]
    \centering\includegraphics[width=0.95\columnwidth]{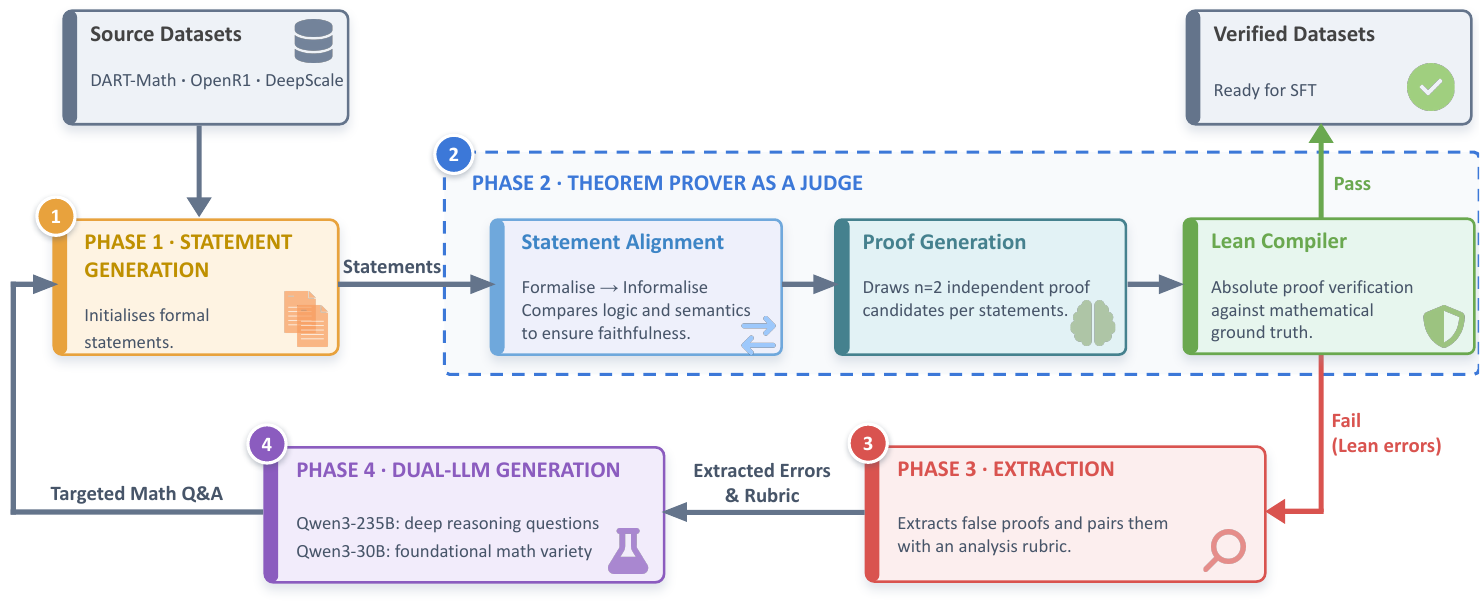}
    \caption{
    The seed synthetic data generation pipeline for \emph{easy} and \emph{medium tier} problems.}
    \label{fig:synthetic-data-pipeline}
\end{figure}

\subsection{Synthetic Data Creation}
\label{sec:synthetic_data}

Our training corpus is produced across three difficulty tiers, gated throughout on the Lean type-checker. \emph{Easy} and \emph{medium} instances are autoformalised from general mathematical-reasoning datasets (\S\ref{sec:easy_medium}), while \emph{hard} instances are drawn from competition-level sources (\S\ref{sec:hard}). The tiers are ordered monotonically by construction: \emph{easy} instances are rubric-driven simplifications of medium ones (\S\ref{sec:rubric}); \emph{medium} problems are routine multi-step reasoning; and \emph{hard} problems are competition proofs requiring non-routine techniques absent from general maths corpora. A representative example from each tier is given in Appendix~\ref{sec:difficulty}. This seed corpus delivers quality at modest scale; \S\ref{sec:alf} then scales it via Augmented Lean Formalisation.

\subsubsection{The Easy and Medium Tiers}
\label{sec:easy_medium}
The easy and medium tiers are produced by a three-stage seed pipeline: formal statement synthesis, formal proof verification, and rubric-guided distillation, shown in Figure~\ref{fig:synthetic-data-pipeline}.

\paragraph{Formal statements and formal proofs.}
Our pipeline is inspired by the theorem prover as a judge pipeline~\citep{leang-etal-2025-theorem}, which leverages the deterministic verification of formal proof assistants to filter synthetic training data. While \citet{leang-etal-2025-theorem} operate solely on problems drawn from GSM8K and MATH500 formalised with GPT-4~\citep{hurst2024gpt}, we broaden both the difficulty range and the domain coverage of the resulting corpus by sourcing natural-language problems from \emph{DART-Math-Hard}~\citep{tong2024dartmath}, \emph{DeepScaleR-Preview}~\citep{deepscaler2025}, and \emph{OpenR1-Math}~\citep{openr1}. Each problem is autoformalised into a Lean statement using Goedel-Autoformaliser-v2~\citep{lin2025goedelproverv2scalingformaltheorem}. To mitigate the logical errors introduced by autoformalisation, we then apply \emph{auto-informalisation}~\citep{leang-etal-2025-theorem}: each candidate formal statement is translated back into natural language and aligned with its source problem, and instances exhibiting logical discrepancies are discarded.

For every retained formal statement, we synthesise candidate proofs with Goedel-Prover-v2-32B~\citep{lin2025goedelproverv2scalingformaltheorem}, an open prover that is orders of magnitude smaller than the GPT-4 model used by \citet{leang-etal-2025-theorem}. We draw $n{=}2$ independent proof attempts per statement and accept the (statement, proof) pair into the training corpus if at least one attempt is verified by the Lean type-checker. This deterministic verification step ensures that every retained training example is verified by Lean, yielding a corpus whose correctness is enforced by the proof assistant.

\paragraph{Rubric-guided distillation.}
\label{sec:rubric}
Existing autoformalisers and provers still fail on a substantial fraction of complex mathematical statements~\citep{lin2025goedelproverv2scalingformaltheorem, DBLP:journals/corr/abs-2504-21801, wang2025kiminaproverpreviewlargeformal}. To characterise these failures, we parse the errors shown by the Lean compiler over all rejected instances and additionally annotate a representative subset manually; we then group the observed errors into seven recurring categories:  (i)~\emph{invalid projection or field errors}, where the proof references a non-existent field of a mathematical object; (ii)~\emph{unsolved goals}, where residual goal states remain after proof execution; (iii)~\emph{tactic failures} (e.g.\ \texttt{linarith}, \texttt{simp}, \texttt{rewrite}), arising when algebraic expressions resist automated simplification; (iv)~\emph{type mismatches}, involving domain conflicts such as integers versus natural numbers; (v)~\emph{synthesis failures}, where a required type-class property (e.g.\ finiteness, decidability) cannot be inferred; (vi)~\emph{unknown constant or identifier errors}, typically arising from hallucinated theorem names; and (vii)~\emph{other errors}, covering miscellaneous technical failures. Further explanation and examples of each error are presented in Appendix~\ref{app:analysis_lean}.

This taxonomy forms the basis of a rubric (Appendix~\ref{app:analysis_lean}) that prompts an LLM to generate simplified variants of each failed problem, with each variant targeted at the specific algebraic manipulation or logical step responsible for the failure. Unlike the free-form scaffolded synthesis of prior work~\citep{lin2025goedelproverv2scalingformaltheorem}, which prompts for unconstrained simpler or harder variants, our rubric is conditioned on a fine-grained taxonomy of Lean failure modes, so that each generated variant addresses a concrete cause of failure.
We synthesise two variants per instance with Qwen3-235B and Qwen3-30B~\citep{DBLP:journals/corr/abs-2505-09388}, pairing a high-capacity generator with a smaller model that tends to yield more diverse, simpler and more effective task-specific outputs~\citep{kim-etal-2025-evaluating}.
Applying the rubric yields a $30\%$ relative improvement in autoformalisation success rate; further analysis and decomposition are provided in \S\ref{sec:dataset_decom}.

\subsubsection{The Hard Tier}
\label{sec:hard}

For the hard tier we draw on Big-Math-RL-Verified~\citep{albalak2025bigmathlargescalehighqualitymath}, restricting our selection to its Olympiads, AIME, AMC, \texttt{aops\_forum}, and Level~4/5 MATH subsets. This choice is motivated by the difficulty profile of the easy/medium sources, which skews toward routine problems: DART-Math-Hard is rejection-sampled from GSM8K and MATH; DeepScaleR interleaves AIME and AMC problems with Omni-MATH~\citep{gao2024omnimathuniversalolympiadlevel} and routine contest preparation; and OpenR1-Math is distilled from NuminaMath-CoT~\citep{numina_math_datasets}, which is itself dominated by lower-difficulty sources such as \texttt{cn\_k12}. Restricting the hard tier to the subsets above therefore concentrates it on problems that demand multi-step, proof-oriented reasoning over advanced number theory, combinatorics, inequalities, and Euclidean geometry. When an exact or near-duplicate appears in both easy/medium and hard sources, we assign it to the hard tier and remove it from easy/medium to avoid cross-tier contamination.

From each retained problem we synthesise additional question variants with Qwen3.6-27B~\citep{qwen3.6-27b} and pass the variants through the same formal-statement and formal-proof pipeline as~\citet{leang-etal-2025-theorem} (step 2 in Figure~\ref{fig:synthetic-data-pipeline}). Only the synthesised variants enter the supervised fine-tuning corpus; the original Big-Math hard problems are held out and reserved exclusively for reinforcement learning. As the hard tier is substantially more challenging, we synthesise $n=4$ variants per retained problem (versus $n=2$ for easy/medium) so that the pipeline retains enough verifiable instances after Lean filtering. We deliberately keep these originals at full difficulty so that they retain their value both as a challenging on-policy reward signal and as an unleaked probe of generalisation; we therefore do \emph{not} apply rubric-guided distillation to the hard tier, since its simplification step would convert hard-source failures into local subproblems and leak hard-problem structure into the supervised corpus.

\begin{figure}[t]
    \centering\includegraphics[width=0.99\columnwidth]{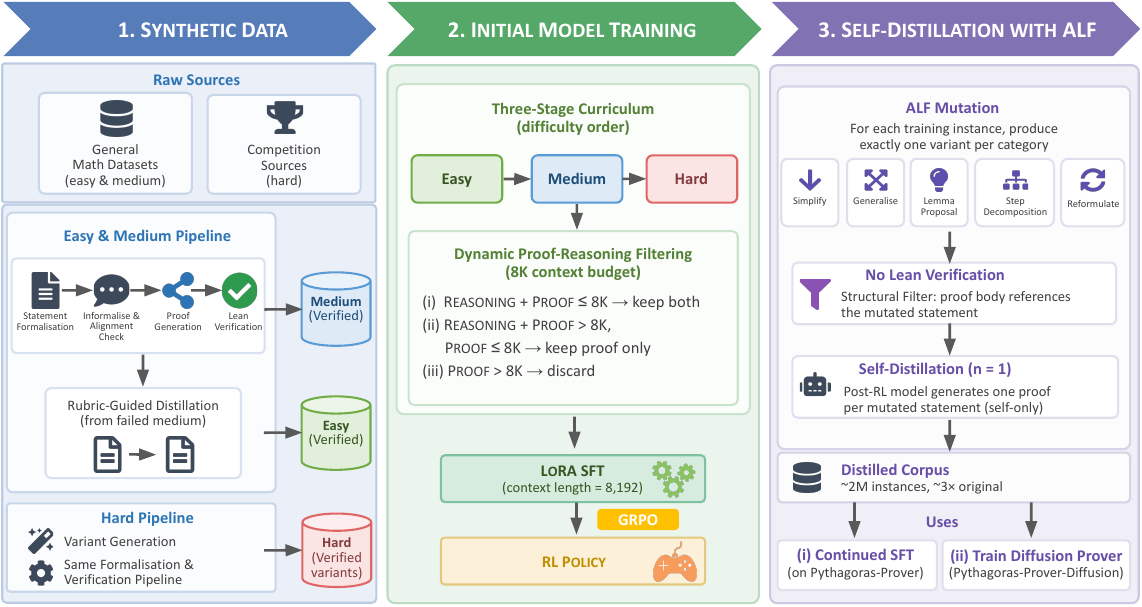}
    \caption{Overview of \method pipeline.}
    \label{fig:pythagoras_pipeline}
\end{figure}

\subsection{Training Algorithm}
\label{sec:training_algorithm}
We train \method in three stages: supervised fine-tuning on the seed corpus under a difficulty-ordered curriculum, reinforcement learning on held-out hard problems, and a final continued-SFT stage on the ALF self-distillation corpus of \S\ref{sec:alf}.

\paragraph{Supervised fine-tuning and a three-stage curriculum.}
We perform supervised fine-tuning with LoRA~\citep{DBLP:conf/iclr/HuSWALWWC22} under a context length of $8{,}192$ tokens, chosen to balance training stability against computational cost. To our knowledge, \method is the first prover to reach state-of-the-art performance using parameter-efficient supervised fine-tuning, whereas prior open provers rely on full-parameter training~\citep{lin2025goedelproverv2scalingformaltheorem, DBLP:journals/corr/abs-2504-21801}. We order training by difficulty in a three-stage curriculum that mirrors the tiers of \S\ref{sec:synthetic_data}, training first on easy instances, then medium, and finally hard. Within each stage, a fixed 8K budget is nonetheless restrictive for theorem proving, where reasoning chains are long and discarding them outright degrades performance. We therefore adopt a \emph{dynamic proof-reasoning filtering} scheme governed by three cases: (i)~if the reasoning chain and formal proof together fit within the budget, we train on the complete sequence; (ii)~if their combined length exceeds the budget, we drop the reasoning chain and retain only the formal proof; and (iii)~if the formal proof alone exceeds the budget, the instance is discarded. This scheme improves substantially over both naive truncation and naive instance filtering, recovering performance comparable to full-context fine-tuning at a fraction of the cost (Appendix~\ref{sec:dynamic_filtering_ablation}). More training details are provided in Appendix~\ref{app:training-details}.

\paragraph{Reinforcement learning.}
For the reinforcement-learning stage we aggregate three sources: Goedel instances generated without self-correction~\citep{lin2025goedelproverv2scalingformaltheorem}, the manually annotated subset of NuminaMath-Lean~\citep{kimina_prover_2025}, and the original Big-Math hard problems held out from \S\ref{sec:hard}. We train with GRPO~\citep{DBLP:journals/corr/abs-2402-03300} with a single epoch using a rollout size of $8$ per problem; a rollout receives a reward iff its proof compiles under Lean and aligns with the formal statement, and $0$ otherwise. We apply DAPO-style~\citep{DBLP:journals/corr/abs-2503-14476} dynamic filtering, retaining a problem only when its number of successful rollouts lies in $\{1,\ldots,5\}$.
We further remove the KL-divergence penalty to encourage exploration. Unlike the SFT stage, we use full-parameter fine-tuning here, having found the LoRA-adapted policy unstable under GRPO updates~\citep{qi2025defeating}. We observe that RL yields only modest gains over the curriculum-SFT model. We interpret this not as a shortcoming of RL but as evidence of the quality of our training corpus: because every supervised instance is Lean-verified and presented in difficulty order, the policy already internalises strong proof-search behaviour before RL, leaving limited headroom for on-policy improvement. Appendix~\ref{app:decomposition} quantifies the contribution of each stage.

\subsection{Augmented Lean Formalisation and Self-Distillation}
\label{sec:alf}
The seed corpus of \S\ref{sec:synthetic_data} is bounded in size by the throughput of Lean verification: every retained instance has been gated on the type-checker. To scale beyond this bound, we propose \emph{Augmented Lean Formalisation} (ALF), a structured mutation operator that expands the seed via formal variants without requiring per-instance verification. ALF prompts a dedicated mutation model (Qwen3.6-27B) to emit, for each seed instance, exactly one variant in each of five categories: \emph{simplification}, \emph{generalisation}, \emph{lemma proposal}, \emph{proof-step decomposition}, and \emph{reformulation}. Unlike free-form variant generation at test time, ALF is applied offline across the entire seed corpus, yielding balanced coverage over transformation types.

Crucially, ALF dispenses with formal verification of the mutated instances, replacing it with a cheaper consistency check. The pipeline runs in three steps, each using a distinct, deliberately small model where a model is needed. \emph{(i) Statement mutation:} a dedicated mutation model, Qwen3.6-27B, rewrites each seed statement into one formal variant per category, producing mutated \emph{statements} only (no proofs at this stage). \emph{(ii) Proof self-distillation:} each mutated statement is then populated with a proof by the post-RL \method of \S\ref{sec:training_algorithm}, which generates a single candidate proof ($n{=}1$) per statement; the 4B and 32B models each prove their own mutations. \emph{(iii) Statement-alignment filtering:} in place of Lean verification, we retain a (statement, proof) pair only when the generated proof references its target formal statement.

We stress that step (iii) is a \emph{statement-alignment} check, not a correctness check: it confirms that the generated proof addresses the intended goal, but it never invokes the Lean type-checker and does not certify that the proof is valid. Consequently, unlike the seed corpus, the ALF data is not Lean-verified. Replacing per-instance verification with this single alignment check removes the dominant computational cost of the pipeline, as end-to-end Lean compilation at this scale remains prohibitively expensive. We adopt this choice because prior work shows that unverified self-distilled samples remain beneficial even in domains with an executable correctness signal, such as coding~\citep{DBLP:journals/corr/abs-2604-01193}; whether the same holds for formal proving is an empirical question, which we examine in Appendix~\ref{app:decomposition}. As a check, we Lean-verify a random subset of $2{,}000$ ALF instances and find that $87.8\%$ pass, indicating that quality remains high despite the absence of exhaustive verification.

The filtered generations, together with the verified proofs collected during reinforcement learning, form a corpus of approximately 2M instances, roughly $2.5\times$ the size of the seed, which we use both for a continued-SFT stage of \method (\S\ref{sec:training_algorithm}) and for training the diffusion-based prover (\S\ref{sec:scaling_diffusion}).

We complete training with a continued-SFT stage on this corpus, again using LoRA to bound compute. We found cross-model distillation (for example, a 32B generator paired with a 4B trainee) unstable, so no external or larger teacher is used. The same corpus is also used to train the diffusion-based prover of \S\ref{sec:scaling_diffusion}, so both branches of the \method family share a single self-distillation data recipe.

\subsection{\method-Diffusion}
\label{sec:scaling_diffusion}

As an alternative use of the self-distillation corpus, we train \method-Diffusion, to our knowledge the first diffusion-based theorem prover. We build on a block-diffusion formulation~\citep{DBLP:conf/iclr/ArriolaGCYQHSK25} using the \texttt{dllm} framework~\citep{DBLP:journals/corr/abs-2602-22661}, in which the proof is partitioned into blocks generated autoregressively while the tokens within each block are produced by discrete diffusion. The transition from autoregressive to diffusion language models has been reported to destabilise as the target length grows~\citep{DBLP:journals/corr/abs-2510-04146}; we find that the \emph{dynamic proof-reasoning filtering scheme} of \S\ref{sec:training_algorithm} carries over to this regime and keeps training stable on long Lean proofs. The remainder of this section describes our central modification to the within-block diffusion: \emph{tactic-based masking}, which corrupts each proof at the granularity of complete Lean tactics so that the denoising objective is aligned with the discrete reasoning steps the prover must commit to at inference time.

\paragraph{Tactic-based masking.}
We use masked diffusion rather than autoregressive decoding because diffusion models can fill in sequence positions in any order, revisit earlier commitments once later context disambiguates them, and commit several positions per forward pass with a more efficient cost \citep{sahoo2024simple, DBLP:journals/corr/abs-2502-09992}; in our case the unit of corruption is a Lean tactic rather than a single token. We adopt the discrete masked diffusion formulation of \citet{DBLP:journals/corr/abs-2502-09992} under the linear schedule $\alpha_t = 1 - t$, so that the diffusion time $t \in [0,1]$ coincides with the per-unit mask probability. Let $\mathbf{x}_0 = (x_0^1, \ldots, x_0^L)$ denote a clean Lean proof of length $L$, $\mathbf{x}_t = (x_t^1, \ldots, x_t^L)$ its corrupted counterpart containing the mask symbol $[\textsc{m}]$, and $p_\theta(\cdot \mid \mathbf{x}_t)$ the denoiser that predicts the clean token at every masked position. The theorem statement (header, goal, \texttt{:=by}) is treated as a fixed input block on which the denoiser always conditions, and corruption is applied only to positions within the proof body; the diffusion time $t$ is resampled independently for each training sequence.
Random per-token masking treats every position as exchangeable, which is mismatched with Lean proofs: the natural unit of reasoning is a complete \emph{tactic}, and we want the denoiser to recover whole tactics rather than isolated tokens that can be guessed from sibling tokens within the same tactic. We therefore 
change only the unit of corruption. Let $\mathcal{T}(\mathbf{x}_0) = (\tau_1, \ldots, \tau_K)$ be the tactic spans extracted from $\mathbf{x}_0$ by a Lean parser (examples for such a span would be \mintinline{lean}|intro x y z| or 
\mintinline{lean}|have hnum : x * y * z * (x + y + z) = 36 := by|), each $\tau_k \subseteq \{1, \ldots, L\}$ a contiguous range of token indices. The residual positions $S = \{1, \ldots, L\} \setminus \bigcup_k \tau_k$ are structural (the \texttt{by} keyword, indentation, the theorem header) and are never masked: $x_t^i = x_0^i$ for all $i \in S$. The remaining positions are corrupted span-wise:\footnote{For a symbol $a$, a sequence $\mathbf{x}$ and a set of integers $b$, we denote by $a^{|b|}$ the symbol repeated $|b|$ times and by $\mathbf{x}^b$ the subsequence indexed by elements of $b$.}
\begin{equation}
    \mathbf{x}_t^{\tau_k} \;\sim\;
    \begin{cases}
        [\textsc{m}]^{|\tau_k|} & \text{with probability\ } t, \\[2pt]
        \mathbf{x}_0^{\tau_k}            & \text{with probability\ } 1 - t,
    \end{cases}
    \qquad k = 1, \ldots, K,
    \label{eq:tac-forward}
\end{equation}
independently across tactics. A whole tactic is therefore either masked in full or kept intact; partial masking inside a tactic does not occur. The denoiser remains a per-token model, and the training objective replaces the per-position sum of the standard MDLM cross-entropy with a per-span sum that aggregates the per-token log-probabilities produced by $p_\theta$:
\begin{equation}
    \mathcal{L}_{\mathrm{tac}}(\theta) \;=\; -\,\mathbb{E}_{t \sim \mathcal{U}[0,1],\, \mathbf{x}_0,\, \mathbf{x}_t}\!\left[\, \frac{1}{t} \sum_{k=1}^{K} \mathbf{1}\!\left[\mathbf{x}_t^{\tau_k} = [\textsc{m}]^{|\tau_k|}\right] \sum_{i \in \tau_k} \log p_\theta\!\left(x_0^i \,\middle|\, \mathbf{x}_t\right) \right],
    \label{eq:tac-loss}
\end{equation}
in which the $1/t$ prefactor compensates the expected mask count so that every noise level contributes equally in expectation to the loss. Setting $\tau_k = \{k\}$ for every $k$ ($K=L$ in that case) reduces Eq.~\eqref{eq:tac-loss} to the per-token MDLM objective of \citet{sahoo2024simple}, so tactic-level masking is a strict generalisation of token-level masking. Because an entire tactic is masked or retained as a unit, the model cannot exploit sibling tokens within the same tactic and must recover the whole proof step from the surrounding context; the expected number of masked positions per sample, $t(L - |S|)$, scales linearly in $t$, so the $1/t$ weighting yields a constant-in-expectation contribution across noise levels for all tokens within a single example in a batch.

\section{Experimental Setup}
In this section, we describe the experimental setup used to evaluate \method, covering the benchmarks~\S\ref{subsec:benchmark} and the evaluation protocol~\S\ref{subsec:eval}. All design choices are kept consistent across runs unless explicitly stated.

\subsection{Benchmarks}
\label{subsec:benchmark}
We evaluate on three Lean theorem-proving benchmarks spanning a range of difficulty and topical coverage: MiniF2F, PutnamBench, and \emph{MiniF2F-ALF}, a companion benchmark we introduce that reuses the ALF mutation operator of \S\ref{sec:alf} as a perturbation on the MiniF2F test set.

\paragraph{MiniF2F.} MiniF2F~\citep{DBLP:journals/corr/abs-2109-00110} comprises 488 problem statements in Lean (244 validation and 244 test) drawn from high-school level competitions, including the AMC, AIME, and the International Mathematical Olympiad (IMO). We adopt the Kimina-revised release~\citep{wang2025kiminaproverpreviewlargeformal} and report results on MiniF2F-Test, in which several erroneous statements from the original release have been corrected.

\paragraph{PutnamBench.} 
PutnamBench~\citep{tsoukalas2024putnambenchevaluatingneuraltheoremprovers} targets college-level mathematics through 672 problems sourced from the William Lowell Putnam Mathematical Competition (1962--2023), covering algebra, analysis, number theory, geometry, combinatorics, probability, and set theory.

\paragraph{MiniF2F-ALF.}

MiniF2F-ALF is a mutated variant of MiniF2F-Test, constructed by applying the ALF mutation scheme of \S\ref{sec:alf} to each of the 244 test statements, together with further numerical and variable perturbation inspired by \citet{DBLP:conf/iclr/MirzadehASTBF25}. It plays a dual role. First, it serves as a \emph{contamination-sensitive perturbation probe}: models that rely heavily on exact benchmark recall should degrade under controlled statement mutations, whereas models with more transferable proof behaviour should retain more performance. Second, it serves as a \emph{transfer probe}, testing whether models trained on ALF-augmented data generalise to ALF-mutated test data. ALF defines five mutation operators, but some produce variants that remain nearly identical to the source statement and provide little signal for either purpose. We therefore generate five candidate mutations per problem with Codex (GPT-5.5)~\citep{DBLP:journals/corr/abs-2601-03267} and retain the two most divergent under cosine distance in the \texttt{Alibaba-NLP/gte-Qwen2-7B-instruct} embedding space, yielding a 488-statement benchmark. We verify that every retained statement is a well-formed Lean theorem and further judge well-formedness using Claude Code (Claude-Opus-4.5~\citep{anthropic2025claude45systemcard}). Further details, including the perturbation procedure, are provided in Appendix~\ref{app:minif2f_settings}.

\subsection{Evaluation}
\label{subsec:eval}

All experimental results are evaluated with Lean 4.9.0-rc1, using a similar evaluation environment as previous work~\citep{lin2025goedelproverv2scalingformaltheorem, DBLP:journals/corr/abs-2504-21801}.
The model's maximum generation length is set to 30{,}000 tokens. A proof attempt is judged correct only if (i) it compiles under Lean with no errors and contains no \texttt{sorry}, \texttt{admit}, or otherwise unproved goals, and (ii) the target formal statement appears verbatim in the generated output, ensuring that the model proves the stated goal rather than a rewritten or weaker one. We report pass@$N$, the fraction of problems for which at least one of $N$ independently sampled attempts is judged correct. 

\section{Experimental Results}
\label{sec:results}

\begin{table}[t]
\centering
\caption{Pass@$N$ (\%) on MiniF2F-Test. ``Best ($N$)'' represents each method's highest reported pass rate, with the corresponding sampling budget in parentheses. $^\dagger$ marks Kimina-Prover variants, and the highest overall result is highlighted in bold. Results in the upper section are copied from the respective papers. A detailed comparison is provided in Table~\ref{tab:budget_performance_full}.}
\label{tab:budget_performance}
\small
\begin{tabular}{@{}l r r r r@{}}
\toprule
\textbf{Method} & \textbf{\#Params} & \textbf{Pass@32} & \textbf{Pass@1024} & \textbf{Best ($N$)} \\
\midrule
Goedel-Prover-SFT~\citep{DBLP:journals/corr/abs-2502-07640} & 7B & 57.6 & -- & 62.7 (3200) \\
STP~\citep{DBLP:conf/icml/Dong025a} & 7B & -- & -- & 67.6 (25600) \\
Kimina-Prover-Preview-72B~\citep{wang2025kiminaproverpreviewlargeformal} & 72B & 68.85 & -- & 80.74 (8192) \\
DeepSeek-Prover-V2-7B~\citep{DBLP:journals/corr/abs-2504-21801} & 7B & 75.6 & -- & 82.0 (8192) \\
DeepSeek-Prover-V2-671B & 671B & 82.4 & -- & 88.9 (8192) \\
Kimina-Prover-8B-Distill$^\dagger$~\citep{wang2025kiminaproverpreviewlargeformal} & 8B & 77.86 & -- & -- \\
Kimina-Prover-70B$^\dagger$ & 70B & 84.0 & 87.7 & 92.2 (TTRL)\textsuperscript{*} \\
Goedel-Prover-V2-8B~\citep{lin2025goedelproverv2scalingformaltheorem} & 8B & 84.6 & 87.9 & 90.2 (8192) \\
\quad\textit{(+Self Correction)} & 8B & 86.7 & 89.3& -- \\
Goedel-Prover-V2-32B & 32B & 88.1 & 91.8 & 92.2 (8192) \\
\quad\textit{(+Self Correction)} & 32B & 90.4 & 92.6 & -- \\
\midrule
\rowcolor{gray!15}\textbf{\method-4B} & 4B & 86.1 & 88.1 & 89.8 (2048) \\
\rowcolor{gray!15}\textbf{\method-32B} & 32B & 89.8 & {92.6} & \textbf{93.0} (2048) \\
\bottomrule
\end{tabular}

\vspace{0.5em}
\begin{minipage}{\textwidth}
\footnotesize
\textsuperscript{*}Refers to test-time reinforcement learning.
\end{minipage}

\end{table}

\begin{wrapfigure}[17]{r}{0.4\textwidth}
    \centering
     \includegraphics[width=0.4\textwidth]{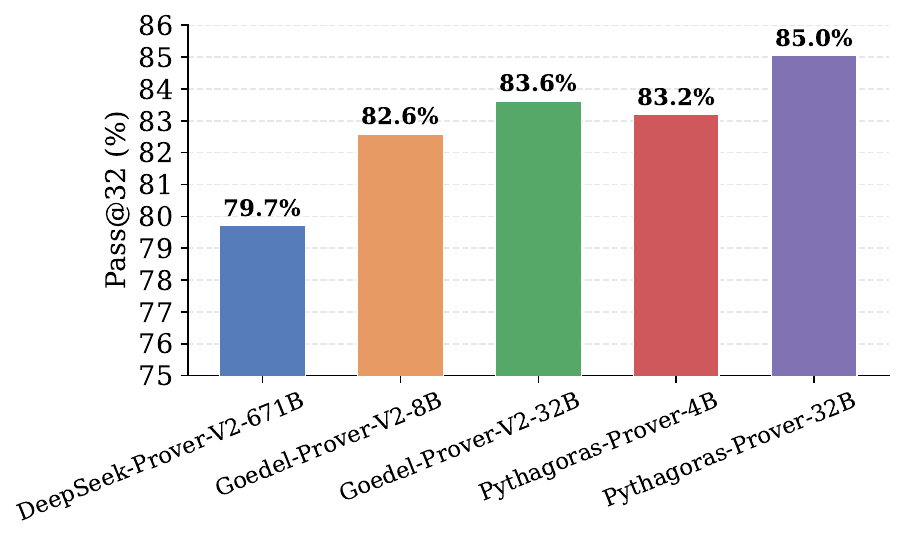}
    \caption{Performance of current state-of-the-art provers on MiniF2F-ALF. As MiniF2F-ALF is introduced in this work, all results are evaluated by us under a unified setup.}
    \label{fig:results-minif2f-mutated}
\end{wrapfigure}

The evaluation results on MiniF2F-Test, PutnamBench, and MiniF2F-ALF are shown in Tables~\ref{tab:budget_performance}--\ref{tab:putnambench} and Figure~\ref{fig:results-minif2f-mutated}, respectively. Below we summarise and discuss the results: state-of-the-art accuracy at modest scale, the strongest open-source result on PutnamBench under matched effective compute, robustness to statement-level perturbation under MiniF2F-ALF, and a first viable diffusion-based theorem prover trained on the same corpus.

\paragraph{State-of-the-art performance at modest scale.}
\method-32B achieves the highest reported pass rate on MiniF2F-Test, $\mathbf{93.03\%}$ at pass@2048, exceeding the best prior open-source result (Goedel-Prover-V2-32B, $92.2\%$ at pass@8192) while using a quarter of the sampling budget, and doing so without inference-time self-correction or test-time reinforcement learning. Among methods that likewise forgo self-correction, \method-32B leads at every budget we evaluate: at pass@32 it attains $89.8\%$, ahead of Goedel-Prover-V2-32B ($88.1\%$), DeepSeek-Prover-V2-671B ($82.4\%$, roughly $20\times$ larger), and Kimina-Prover-70B ($84.0\%$). It further matches the self-correction-augmented Goedel-Prover-V2-32B at pass@1024 ($92.6\%$ vs.\ $92.6\%$) and surpasses it at pass@2048, indicating that the proof-search behaviour internalised during training largely obviates multi-round repair.

\paragraph{\method-4B outperforms substantially larger baselines.}
The efficiency of our recipe is clearest at the 4B scale. Under a matched pass@32 budget, \method-4B attains $86.1\%$, exceeding DeepSeek-Prover-V2-671B ($82.4\%$) by $3.6$ points despite being roughly $167\times$ smaller, and outperforming Kimina-Prover-70B ($84.0\%$), Goedel-Prover-V2-8B ($84.6\%$), and DeepSeek-Prover-V2-7B ($75.6\%$). The advantage persists with budget: at pass@2048 \method-4B reaches $89.8\%$, exceeding DeepSeek-Prover-V2-671B's pass@8192 result ($88.9\%$) at a quarter of the budget. Against the 8B Goedel-Prover-V2, \method-4B leads at every shared budget without self-correction ($88.1\%$ vs.\ $87.9\%$ at pass@1024) and stays within $1.2$ points even of its self-corrected variant, at half the parameters. The training recipe thus transfers cleanly to the small-model regime, where billion-scale baselines are impractical.

\paragraph{PutnamBench: restart sampling reaches the strongest open-source result.}

On PutnamBench (Table~\ref{tab:putnambench}), the comparison is more nuanced. Indeed, self-correction is more sample-efficient at small budgets: Goedel-Prover-V2 with self-correction solves $57$ problems at pass@32, compared with \method's $48$. However, \method uses independent \emph{restart} sampling, generating fresh complete proofs and verifying each with Lean. This strategy scales well with the sampling budget: \method solves $59$ problems at pass@64 and $93$ at pass@2048, achieving the \emph{strongest open-source} result on the leaderboard. This exceeds Goedel-Prover-V2's $86$ solved problems at pass@184 with self-correction by $7$ problems, and nearly doubles DeepSeek-Prover-V2's $47$ at pass@1024. This suggests that, although self-correction can be effective at low budgets, it may become brittle when the initial proof attempt contains an early logical error, since subsequent correction rounds often inherit the same mistake and fail to recover (Appendix~\ref{app:self-correction-failure}). Restart sampling avoids this failure mode by starting each attempt from scratch. As we show next, it is also competitive in terms of \emph{effective} decoding compute.

\begin{table}[t]
\centering
\caption{PutnamBench leaderboard (problems solved out of 672). \method solves 93 problems at pass@2048, the best open-source result, surpassing the previous best (Goedel-Prover-V2, 86 at pass@184 with self-correction).\textsuperscript{*}}
\label{tab:putnambench}
\begin{tabular}{cl r c l}
\toprule
\# & \textbf{Model} & \textbf{Num-solved} & \textbf{Open-source} & \textbf{Compute} \\
\midrule
\rowcolor{gray!15}1 & \textbf{\method} & 93 & \ding{51} & pass@2048 \\
\rowcolor{gray!15}1 & \textbf{\method} & 59 & \ding{51} & pass@64 \\
\rowcolor{gray!15}1 & \textbf{\method} & 48 & \ding{51} & pass@32 \\

2 & Goedel-Prover-V2 (self-correction mode) & 86 & \ding{51} & pass@184 \\
2 & Goedel-Prover-V2 (self-correction mode) & 57 & \ding{51} & pass@32 \\
2 & Goedel-Prover-V2 & 43 & \ding{51} & pass@32 \\
3 & DeepSeek-Prover-V2 & 47 & \ding{51} & pass@1024 \\
3 & DeepSeek-Prover-V2 & 22 & \ding{51} & pass@32 \\
4 & DSP+ & 23 & \ding{51} & pass@128 \\
5 & Bourbaki & 14 & \ding{51} & pass@512 \\
6 & Kimina-Prover-7B-Distill & 10 & \ding{51} & pass@192 \\
7 & Self-play Theorem Prover & 8 & \ding{51} & pass@3200 \\
8 & Goedel-Prover-SFT & 7 & \ding{51} & pass@512 \\
9 & ABEL & 7 & \ding{55} & pass@596 \\
\bottomrule
\end{tabular}

\vspace{0.5em}
\begin{minipage}{\textwidth}
\footnotesize
\textsuperscript{*} Seed Prover~\citep{chen2025seed}, successfully solved 331 problems on PutnamBench. However, the prover is closed-source, and the computational budget at test time remains unclear.
\end{minipage}
\end{table}

Raw generated-token counts do not fully capture the decoding cost of self-correction. With KV caching, each newly generated token still attends to the entire prefix available at that step. Hence, under dense attention the work of generating $a$ tokens from a context of length $m$ grows super-linearly. To capture this effect, we define the \emph{effective token complexity}, a worst-case proxy for cumulative attention work:
\[
\mathrm{ETC}(m,a) \;=\; m a \;+\; \frac{a(a+1)}{2},
\]
where $ma$ is the cost of attending to the fixed input at every step and $a(a{+}1)/2$ the cost of the growing generated prefix. ETC counts cumulative query--key interaction under an idealised dense-attention model; it is not a wall-clock measurement, and Appendix~\ref{app:etc} states the assumptions and limitations in full. We note that ETC is size-agnostic: it counts query–key interactions without reference to model width, and so neither credits nor penalises \method for its smaller parameter count. We therefore do not read the matched-ETC comparison as a model-size efficiency result; it reflects only the difference in how much context each attempt conditions on: a short prompt for restart sampling versus long accumulated histories for self-correction. Self-correction and restarting differ sharply under this measure: later self-correction rounds condition on long accumulated contexts, whereas each restart conditions only on a short prompt.
Table~\ref{tab:etc-comparison} reports the relevant token statistics. Goedel-Prover-V2 runs three self-correction rounds at pass@184, with generation lengths of $18.4$K, $21.3$K, and $23.0$K and roughly $15$K of additional proof context across rounds two and three, giving an estimated $\mathrm{ETC}\approx1.95\times10^{11}$ (full derivation in Appendix~\ref{app:etc}). \method with $1024$ independent restarts conditioned on a short ($\approx285$-token) prompt with $\approx18.7$K generated tokens each, for $\mathrm{ETC}\approx1.85\times10^{11}$. At a slightly lower ETC, \method solves $88$ problems versus Goedel-Prover-V2's $86$, so restart sampling matches or improves on self-correction without incurring the decoding cost of long correction histories.

\begin{table}[t]
\centering
\caption{Average input and generated token counts used for the effective token complexity (ETC) estimate. Goedel-Prover decodes three self-correction rounds per attempt; \method draws independent restarts from a short input context.}
\label{tab:etc-comparison}
\resizebox{0.7\textwidth}{!}{%
\begin{tabular}{lcrr}
\toprule
\textbf{Method} & \textbf{Pass@} & \textbf{Input context (tokens)} & \textbf{Avg.\ generated tokens} \\
\midrule
Goedel-Prover, round 1 & 184  & $284.88$    & $18{,}367.83$ \\
Goedel-Prover, round 2 & 184  & $7{,}838.74$ & $21{,}638.70$ \\
Goedel-Prover, round 3 & 184  & $8{,}309.06$ & $23{,}835.80$ \\
\midrule
\method restart        & 1024 & $284.88$    & $18{,}741.80$ \\
\bottomrule
\end{tabular}%
}
\end{table}

\paragraph{\method is robust under statement-level perturbation.}
To probe robustness, we re-evaluate DeepSeek-Prover-V2-671B, Goedel-Prover-V2-8B\slash{}32B, and \method-4B\slash{}32B on {MiniF2F-ALF}, applying our pass criterion uniformly. Figure~\ref{fig:results-minif2f-mutated} shows that every model loses accuracy  relative  to MiniF2F,
 confirming that ALF mutation induces
 a non-trivial distribution shift. 
\method-32B retains the highest absolute pass rate of any 
evaluated model ($85.0\%$); \method-4B ($83.2\%$) is almost on par with the $8\times$ larger
 Goedel-Prover-V2-32B ($83.6\%$) and degrades less under perturbation than that model does ($2.9$ vs.\ $4.5$ points). Since our training corpus
  is itself ALF-augmented (\S\ref{sec:alf}), this is the empirical signature we would expect: models trained on ALF mutations should retain accuracy under ALF-mutated test data more readily than models that have not seen this kind of structural variation, and the robustness gap we observe relative to Goedel-Prover-V2 is consistent with this prediction.
MiniF2F-ALF therefore plays two roles simultaneously: it acts as a sharper stress test than the saturated original split, preserving the same problem families while exposing failures under nearby formal variants; and it probes whether the augmentation strategy used during training transfers to the evaluation side. Under both readings, these results do not rule out high-level theorem-family overlap, but they suggest that \method is less sensitive than the evaluated baselines to surface-level changes in MiniF2F statements.

\begin{wrapfigure}[16]{r}{0.4\textwidth}
    \centering
    \includegraphics[width=0.4\textwidth]{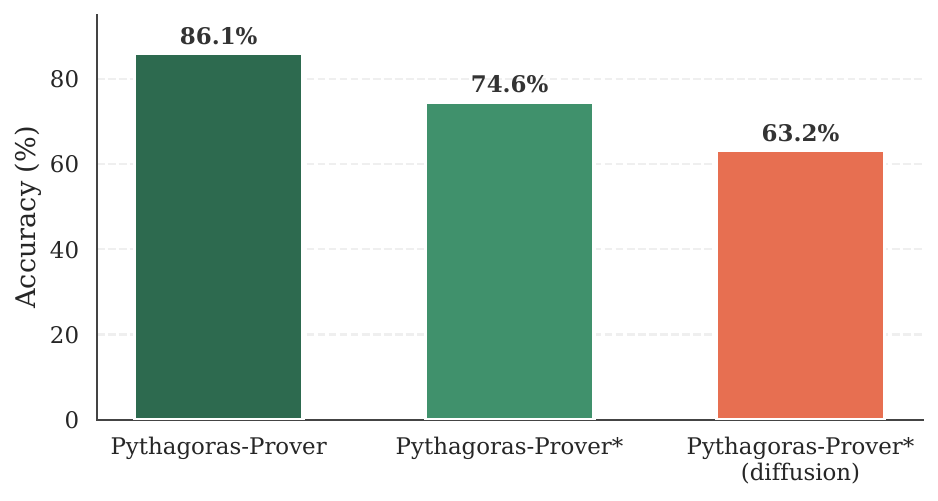}
    \caption{Comparison on the MiniF2F benchmark. $^*$ denotes the setting where training tokens are restricted to 4096 and evaluation is performed solely at 8192 tokens.}
    \label{fig:pythagoras_comparison}
\end{wrapfigure}

\paragraph{\method-Diffusion: the corpus transfers across decoding regimes.}
We additionally train \method-Diffusion on the same self-distillation corpus (\S\ref{sec:alf}) to test whether the data transfers across decoding regimes. To our knowledge, \method-Diffusion is the first viable diffusion-based theorem prover; Figure~\ref{fig:pythagoras_comparison} accordingly establishes an initial baseline for this regime. At pass@32 on MiniF2F-Test, \method-Diffusion-4B attains $63.25\%$ with a block size of $32$, demonstrating that diffusion-style iterative refinement is a viable decoding strategy for Lean proof generation; 
it trails the autoregressive \method-4B ($86.1\%$) by roughly $23$ 
points. We attribute this gap to three constraints that together govern diffusion training for formal mathematics: (i) the random-masking objective, which provides a sparser supervisory signal than next-token prediction; (ii) the limited stable context length of current diffusion language models~\citep{DBLP:journals/corr/abs-2508-15487, DBLP:journals/corr/abs-2602-12586, DBLP:journals/corr/abs-2502-09992} and (iii) the scarcity of diffusion-suitable Lean training data that remain informative while fitting these context limits. We address the first via tactic-based masking (\S\ref{sec:scaling_diffusion}) and the third via the ALF self-distillation corpus (\S\ref{sec:alf}); the second remains binding. Under the autoregressive objective, \method-4B converges stably at the $8{,}192$-token context used throughout this work, whereas the same setting destabilises diffusion training, producing sustained high gradient norms well past warm-up (Figure~\ref{fig:diff_gradnorm}); reducing the context budget to $4{,}096$ tokens recovers stable convergence. To isolate this effect, we re-train the autoregressive \method-4B at the matched $4{,}096$-token context and observe a large performance drop, indicating that a clear bottleneck is the \emph{context length forced by the diffusion regime}. A residual gap in favour of the autoregressive variant remains, which we attribute to the relative maturity of autoregressive foundation models: the Qwen backbone has been pretrained and post-trained far more extensively under the AR objective than under any diffusion objective, so the AR initialisation we fine-tune from is itself a stronger prior for Lean than its diffusion counterpart. Since competitive provers rely on test-time scaling~\citep{chen2025seed, lin2025goedelproverv2scalingformaltheorem, DBLP:journals/corr/abs-2509-22819}, where long reasoning chains routinely exceeding $4{,}096$ tokens are essential to competitive pass rates, the context limit binds tightly today and caps the proofs \method-Diffusion can produce. We therefore present these results as an existence proof rather than a finalised performance number: \method-Diffusion is, to our knowledge, the first demonstration that a diffusion language model can be trained from our corpus to verifiably solve Lean theorems at non-trivial rates. Our belief is that given long enough context window, diffusion models have the potential to capture long-range dependencies that exist in mathematical formal statements and proofs, ranging from simple syntactic dependencies such as parenthetical agreement to more complex dependencies such as logical dependencies between clauses. Diffusion models enable random access to token generation during decoding, which can potentially build a proof section by section while following dependency relations, rather than as a linear sequence of tokens.

\section{Analysis}
In this section, we provide a comprehensive analysis of the dataset decomposition~(\S\ref{sec:dataset_decom}), followed by a scaling study on MiniF2F~(\S\ref{sec:scaling_analysis}). We then examine per-instance behaviour on MiniF2F and MiniF2F-ALF~(\S\ref{sec:minif2f_analysis}), and conclude with a comparison between diffusion and autoregressive provers~(\S\ref{sec:dllm_vs_ar}).

\subsection{Dataset Decomposition}
\label{sec:dataset_decom}

We present a detailed analysis of the Lean-verified seed corpus underlying our training pipeline. The seed is constructed in three difficulty tiers (\S\ref{sec:synthetic_data}), produced by two complementary mechanisms: an initial autoformalisation pass on natural-language problems from existing mathematical-reasoning datasets, which yields the medium and hard tiers (\S\ref{sec:easy_medium}, \S\ref{sec:hard}); and a rubric-guided distillation stage (\S\ref{sec:rubric}) that simplifies failed medium instances to produce the easy tier. The ALF augmentation of \S\ref{sec:alf} is layered on top of this seed and is treated separately.

\begin{figure}[t]
    \centering

    \begin{subfigure}[t]{0.49\textwidth}
        \centering
        \includegraphics[width=\linewidth]{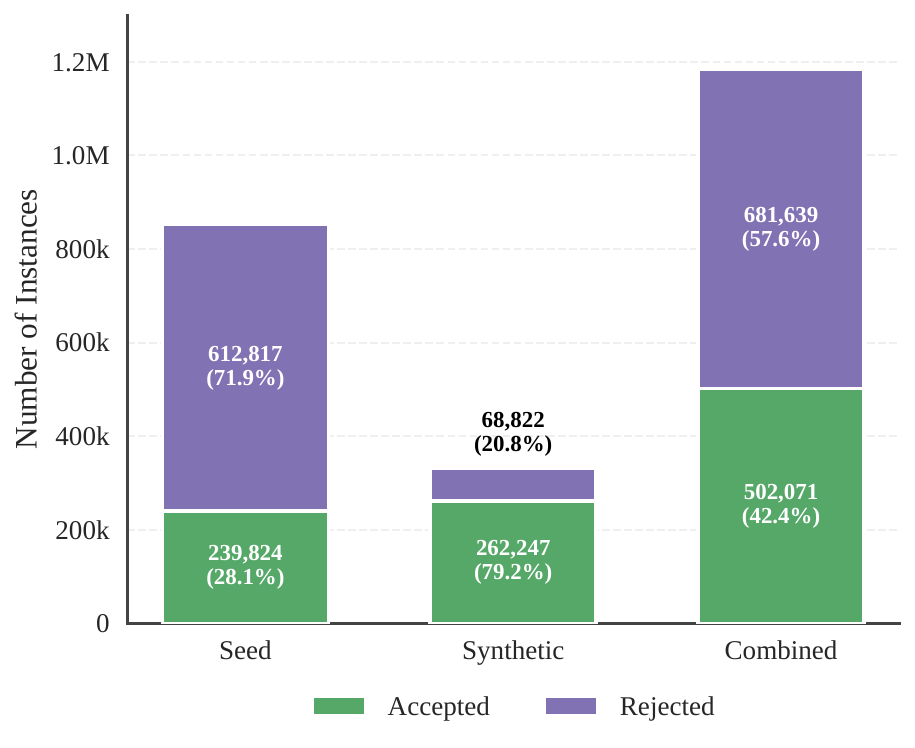}
        \caption{Accepted and rejected instances across seed, synthetic, and combined data.}
        \label{fig:accepted_rejected_totals_ver}
    \end{subfigure}
    \hfill
    \begin{subfigure}[t]{0.49\textwidth}
        \centering
        \includegraphics[width=.62\linewidth]{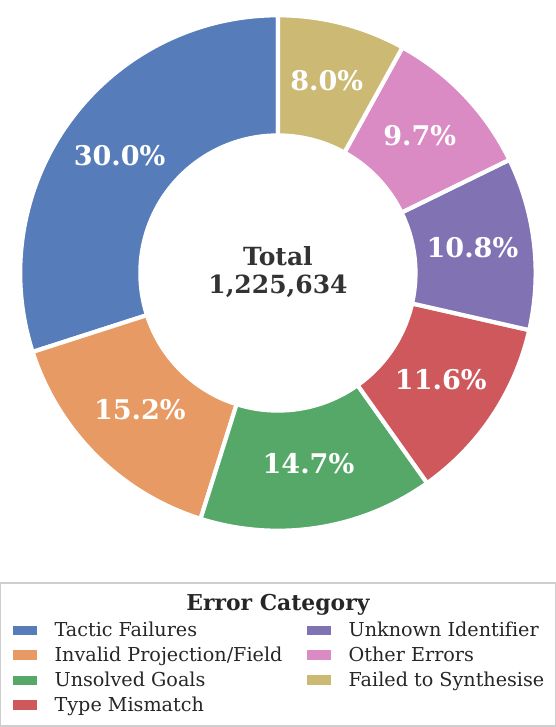}
        \caption{Distribution of error categories for rejected pass@2 attempts.}
        \label{fig:error_distribution_donut}
    \end{subfigure}

    \caption{Data filtering and error analysis. Left: accepted and rejected instances across data sources. Right: distribution of error categories among rejected attempts.}
    \label{fig:data_filtering_error_analysis}
\end{figure}

\subsubsection{The Easy and Medium Tiers}
\label{sec:easy_medium_analysis}

\paragraph{Seed verification (Medium).}
The seed collection from DART-Math-Hard, OpenR1-Math, and DeepScaleR yields $852{,}641$ instances, of which $239{,}824$ ($28.1\%$) are verified by the Lean type-checker and $612{,}817$ ($71.9\%$) are rejected, as shown by the \emph{Seed} bar in Figure~\ref{fig:accepted_rejected_totals_ver}. DART-Math-Hard contributes the majority of verified proofs ($199{,}452$; $83.2\%$ of all verified seeds), reflecting both its larger scale and its emphasis on competition-style problems that tend to be more amenable to automated formalisation. OpenR1-Math and DeepScaleR contribute $28{,}530$ ($11.9\%$) and $11{,}842$ ($4.9\%$) respectively. The substantial gap between DART-Math-Hard and the remaining sources motivates the rubric-guided distillation stage below, which targets the harder instances that initial proof synthesis could not resolve.

\paragraph{Error analysis and rubric-guided synthesis.}
We categorise every type-checker failure on the $612{,}817$ rejected seeds into the seven error classes of \S\ref{sec:rubric}; Figure~\ref{fig:error_distribution_donut} reports the distribution. The dominant failure modes are tactic failures ($30.0\%$, concentrated in \texttt{linarith}, \texttt{simp}, \texttt{ring}, and \texttt{omega}), unsolved goals ($14.7\%$), invalid projection/field errors ($15.2\%$), and type mismatches ($11.6\%$, typically between $\mathbb{N}$ and $\mathbb{Z}$ or related subtypes); unknown-identifier ($10.8\%$), failed-to-synthesise ($8.0\%$), and a long tail of miscellaneous errors ($9.7\%$) account for the remainder. As these categories are not mutually exclusive at the instance level, our rubric addresses each class with targeted simplification rather than a single monolithic correction. From the rejected pool we retain $258{,}648$ instances with actionable error signatures and have Qwen3-30B and Qwen3-235B each produce $n{=}2$ simplified question variants. After near-duplicate filtering, the two models retain $198,641$ and $132,428$ instances respectively, whose union of $331{,}069$ forms the easy synthetic pool.

\paragraph{Synthetic verification (Easy) and final composition.}
As shown by the \emph{Synthetic} bar in Figure~\ref{fig:accepted_rejected_totals_ver}, of the $331{,}069$ candidates entering Lean verification, $262{,}247$ ($79.2\%$) pass and $68{,}822$ ($20.8\%$) are rejected, a $51.1$-point absolute gain over the seed verification rate ($28.1\%$) that is consistent with the rubric's intent: each easy variant isolates the single manipulation responsible for an original failure and is therefore far more amenable to automated proof synthesis. Combined with the seed corpus, the easy/medium dataset comprises $2.04M$ instances with $502{,}071$ verified proofs. Synthetic augmentation alone contributes $262{,}247$ verified instances on top of the $239{,}824$ seed-verified instances, approximately a $2$x increase in verified training data, obtained without any additional human annotation.

\subsubsection{The Hard Tier}
\label{sec:hard_analysis}
The hard tier comprises $328,966$ instances drawn from the hardest subset of our seed sources. Unlike the easy and medium tiers, we do \emph{not} apply rubric-guided synthesis here, since the hardest problems admit no faithful simplification. We instead retain the standard generate-and-verify pipeline and raise the per-problem sampling budget to $n{=}4$ to compensate for the increased difficulty, yielding approximately 1.32M candidate instances before Lean filtering. After Lean verification with $n{=}4$ per instance, $338,683$ instances ($25.74\%$) remain. We read this as evidence that the larger sampling budget offsets the difficulty gap, yielding a hard-tier corpus of comparable density without resorting to simplifications that would alter the underlying problems.

\subsection{Scaling Analysis}
\label{sec:scaling_analysis}
\begin{wrapfigure}[17]{r}{0.5\textwidth}
    \centering
    \vspace{-\intextsep}
    \includegraphics[width=0.5\textwidth]{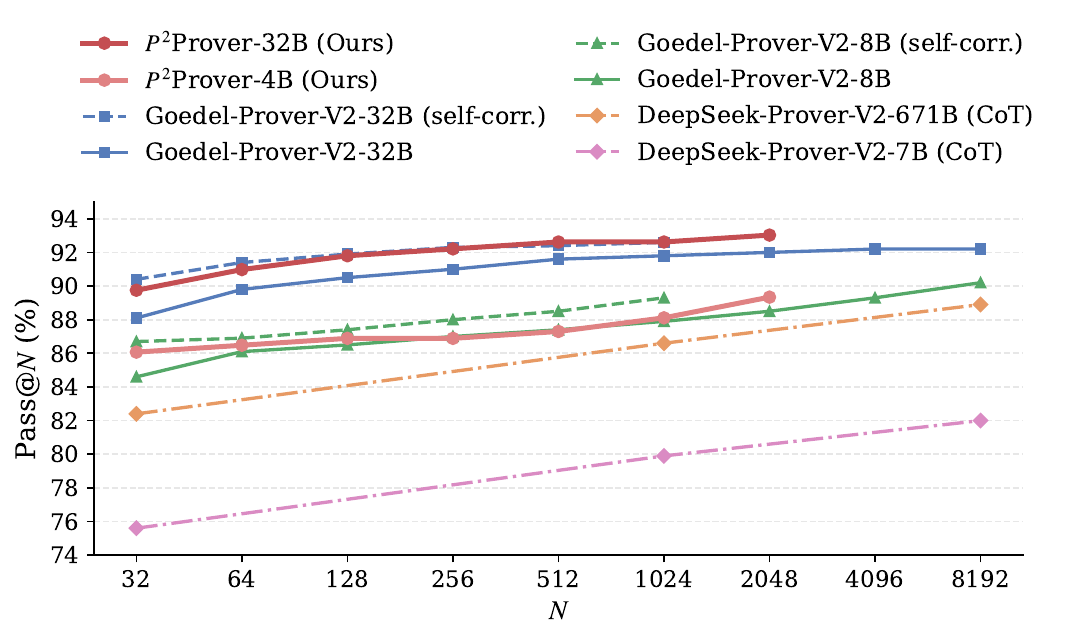}
    \caption{Scaling behaviour on MiniF2F test split. We compare \method with Goedel-Prover-v2, with and without self-correction, and DeepSeek-Prover-v2.}
    \label{fig:pass_n_comparison}
\end{wrapfigure}

Figure~\ref{fig:pass_n_comparison} traces the scaling behaviour of \method (4B and 32B) across budgets $N\in\{32,\ldots,2048\}$ against Goedel-Prover-V2 (8B and 32B, with and without self-correction) and DeepSeek-Prover-V2 (7B and 671B). The advantage is consistent rather than budget-specific. Among methods without self-correction, \method-32B leads at every budget and reaches $93.03\%$ at pass@2048; it crosses above the self-correction-augmented Goedel-Prover-V2-32B from roughly pass@256 onwards, indicating that the proof-search behaviour internalised during training substitutes for iterative verifier-guided correction. 
\method-4B is the more striking case: it surpasses DeepSeek-Prover-V2-671B at every shared budget despite being roughly $167\times$ smaller, reaching $89.8\%$ at pass@2048, which exceeds DeepSeek-Prover-V2-671B's pass@8192 result ($88.9\%$), and it tracks or exceeds the twice-larger Goedel-Prover-V2-8B across the full range. The shape of both curves, high accuracy already at small $N$ with shallow slopes thereafter, shows that \method reaches competitive accuracy with far fewer samples than the baselines, which is the property most relevant to compute-constrained deployment.
These results indicate that \method efficiently internalises proof-search behaviour during training, requiring substantially fewer inference samples to reach competitive accuracy. This is most striking for \method-4B: despite its compact size, it surpasses DeepSeek-Prover-V2-671B across every budget examined throughout the sampling range. Such behaviour highlights that the proposed training recipe transfers cleanly to the small-model regime, yielding a prover whose accuracy--compute profile is well suited to memory- or latency-constrained deployment where billion-scale baselines are impractical. The complementary strength of \method-32B at the upper end of the Pareto frontier confirms that the same recipe scales upward without modification.

\begin{figure}[t]
    \centering\includegraphics[width=\columnwidth]{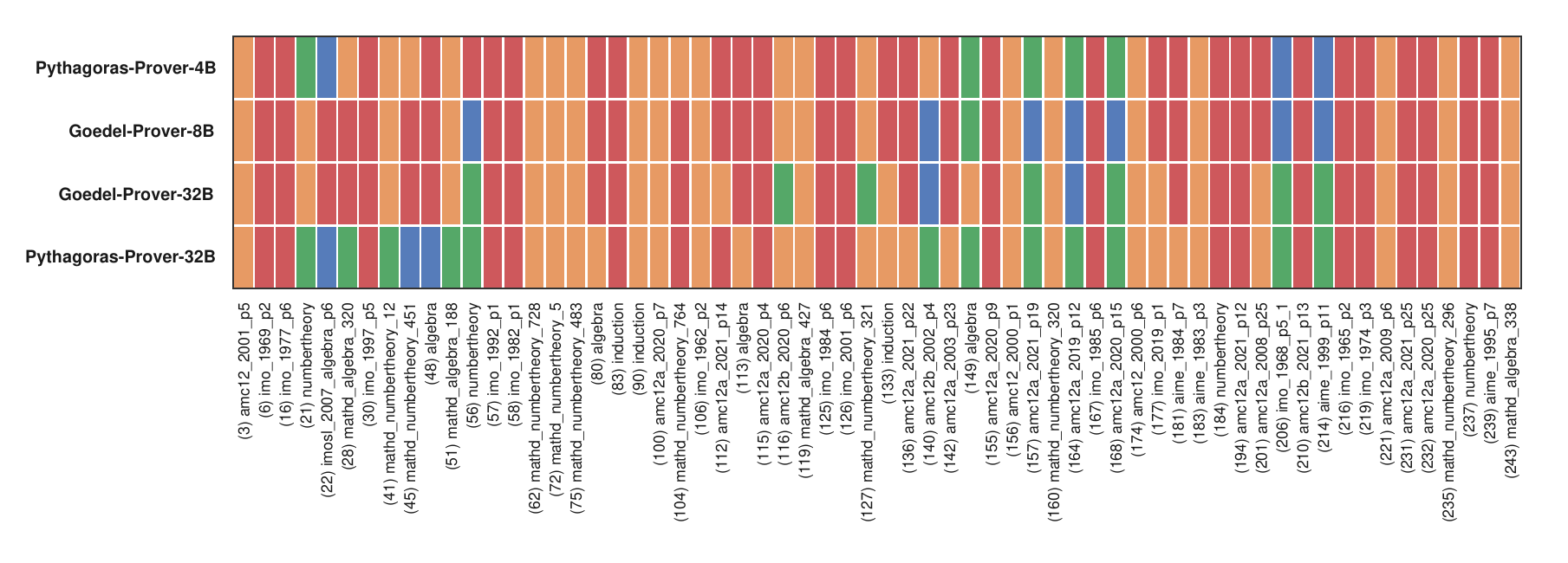}
    \caption{Per-pair outcomes across MiniF2F-Test and its MiniF2F-ALF mutation. {\setlength{\fboxsep}{1.5pt}\colorbox{plotgreen}{\phantom{XX}}}: both solved; {\setlength{\fboxsep}{1.5pt}\colorbox{plotred}{\phantom{XX}}}: both failed; {\setlength{\fboxsep}{1.5pt}\colorbox{plotorange}{\phantom{XX}}}: MiniF2F-Test solved, MiniF2F-ALF failed; {\setlength{\fboxsep}{1.5pt}\colorbox{plotblue}{\phantom{XX}}}: MiniF2F-ALF solved, MiniF2F-Test failed.}
    \label{fig:minif2f_analysed}
\end{figure}

\subsection{Analysis towards MiniF2F}
\label{sec:minif2f_analysis}
\begin{wrapfigure}[15]{r}{0.38\textwidth}
    \centering
    \vspace{-1.0cm}
    \includegraphics[width=0.38\textwidth]{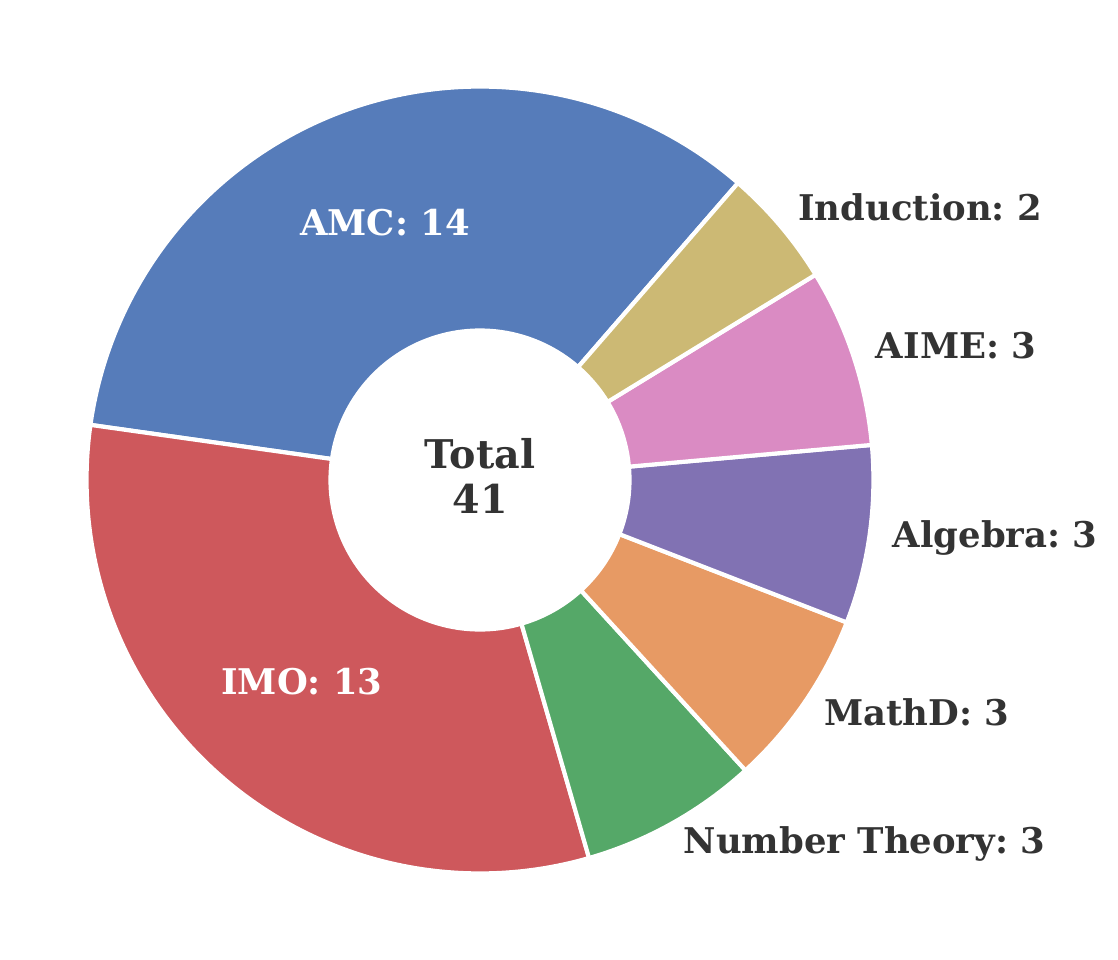}
    \caption{Domain composition of the at-least-one-model-wrong set on original MiniF2F.}
    \label{fig:original_piechart}
\end{wrapfigure}

The headline pass rates of \S\ref{sec:results} already suggest that original MiniF2F is approaching saturation under current Lean provers; the clearer indicator is the size and composition of the residual failure set. Taking \method-4B, \method-32B, Goedel-Prover-V2-8B, and Goedel-Prover-V2-32B\footnote{Goedel-Prover-V2 outcomes throughout this analysis are obtained by our own replication and may differ slightly from those reported by~\citet{lin2025goedelproverv2scalingformaltheorem}.} as a representative panel, ($83.20\%$) of original MiniF2F problems are solved by \emph{all four} models, leaving only $41/244$ ($16.80\%$) on which at least one system fails and $22/244$ ($9.02\%$) on which every system fails. On the bulk of the benchmark, modern provers can no longer be separated; what remains is a small, concentrated tail. That tail is domain-skewed (Figure~\ref{fig:original_piechart}): among 
at-least-one-wrong instances, AMC accounts for $34.15\%$ and IMO for $31.71\%$, and the all-model-wrong subset is more lopsided still, with IMO contributing $50.00\%$ and AMC a further $27.27\%$. Original MiniF2F thus probes a narrow, familiar residual rather than broad reasoning robustness.

\paragraph{MiniF2F-ALF reopens the benchmark and diversifies the failures.}
MiniF2F-ALF holds each problem family fixed and asks whether models remain correct under controlled mutations of the same statement. Under this evaluation the all-model failure rate rises from $9.02\%$ on the original to $13.93\%$ ($68/488$), and the at-least-one-wrong set expands to $18.24\%$ ($89/488$). Critically, the failures \emph{diversify} rather than merely scale: alongside AMC ($30.34\%$) and IMO ($24.72\%$), MathD jumps to $20.22\%$, with Algebra ($7.87\%$), Number Theory ($6.74\%$), Induction ($5.62\%$), and AIME ($4.49\%$) all carrying non-trivial mass. Mutation therefore surfaces brittleness in categories that should not be difficult under robust reasoning but that the saturated original benchmark is too easy to expose. As the same ALF operator drives both training augmentation (\S\ref{sec:alf}) and these test mutations, the redistribution is a sharp diagnostic of where structural generalisation breaks down even in well-trained provers.

\paragraph{Mutation makes strong systems more distinguishable.}
The same effect appears in head-to-head comparisons. On original MiniF2F the systems rarely disagree: \method-4B solves $7$ problems that Goedel-Prover-V2-8B misses while Goedel solves none that \method-4B misses, and \method-32B solves $8$ missed by Goedel-Prover-V2-32B against only $1$ in return. Under ALF the systems disagree on far more instances: \method-4B solves $6$ mutated instances missed by Goedel-Prover-V2-8B with Goedel solving $3$ in return ($9$ disagreements versus $7$ on the original), and \method-32B solves $12$ missed by Goedel-Prover-V2-32B against $5$ ($17$ versus $9$). \method remains favoured in both directions at both scales, but the central point is diagnostic rather than directional: mutation roughly doubles the number of instances on which strong systems can be told apart, recovering discriminative power that the saturated original benchmark has lost. The per-problem transition in Figure~\ref{fig:minif2f_analysed} makes this concrete: the orange region (solved on the original but failed after mutation) isolates exactly the brittleness that high MiniF2F pass rates obscure.

\subsection{Compute Efficiency: \method versus \method-Diffusion}
\label{sec:dllm_vs_ar}
We compare \method-4B and \method-Diffusion-4B under a matched evaluation setting on MiniF2F pass@32: a single $8\times$\,H100 node, batch size $4$, and an identical harness. The two systems are trained on the same self-distillation corpus (\S\ref{sec:alf}) and differ only in decoding regime, so the comparison isolates how the same data behaves under autoregressive versus diffusion sampling. On raw accuracy the autoregressive model is clearly ahead, $86.1\%$ versus $63.3\%$, a gap of $22.7$ points. This comparison, however, counts only per-attempt success and ignores the very different cost of producing each attempt.

Under the harness above, the two systems sustain average generation throughput, calculated as tokens/sec (TPS): $\mathrm{TPS}_{\text{AR}}=4.10$ and $\mathrm{TPS}_{\text{Diff}}=10.56$ output tokens per second per GPU worker, so the diffusion sampler is $2.58\times$ faster. The asymmetry is structural: autoregressive decoding is strictly sequential and becomes expensive on the long completions permitted by \texttt{max\_new\_tokens}$=30{,}000$, whereas the diffusion sampler uses a fixed denoising budget with block-parallel generation and sustains markedly higher throughput on the same hardware.

To weigh accuracy against this cost, we report a throughput-weighted score:
$S = \text{accuracy} \times \text{TPS},$
which is proportional to solved proofs per GPU-second \emph{when the two systems generate a comparable number of tokens per attempt}. This gives $S_{\text{AR}}=0.8607\times4.10=3.53$ and $S_{\text{Diff}}=0.6325\times10.56=6.68$, a $1.89\times$ advantage for the diffusion sampler under this proxy. We read $S$ cautiously: it is a coarse measure whose value at this operating point is dominated by throughput rather than accuracy, so it tracks compute efficiency more than absolute proving capability.

The substantive claim is therefore more limited than the score alone would suggest. For applications that prize per-attempt reliability, the $22.7$-point accuracy gap is decisive and the autoregressive model is preferable. Where throughput or wall-clock latency is the binding constraint, however, \method-Diffusion is the more compute-efficient option at equal hardware, delivering more verified proofs per GPU-second, and it does so as a first-generation, context-limited prover (\S\ref{sec:scaling_diffusion}); we expect this efficiency margin to grow as stable diffusion context lengths increase. We currently read \method-Diffusion as occupying the throughput side of an accuracy--throughput frontier traced by the same training corpus, rather than as competing on the accuracy frontier itself.

\section{Related Work}
\textbf{LLM reasoning.} LLM reasoning abilities have advanced significantly on complex mathematical reasoning tasks~\citep{DBLP:journals/corr/abs-2407-21783}, particularly through chain-of-thought (CoT) reasoning~\citep{wei2022chain}. Large reasoning models (LRMs)~\citep{DBLP:journals/corr/abs-2501-12948, DBLP:journals/corr/abs-2505-09388, DBLP:journals/corr/abs-2503-19786} instead tackle complex problems by generating long CoT traces, and are further enhanced through techniques such as confidence selection~\citep{leang2025picsar}, iterative refinement~\citep{DBLP:journals/corr/abs-2408-03314}, multi-path exploration~\citep{DBLP:conf/icml/GuanZLSSZ0025} and others. It is also well established that test-time scaling improves performance in both general reasoning~\citep{muennighoff2025s1} and formal reasoning~\citep{DBLP:journals/corr/abs-2509-22819}. More recently, diffusion models have proven highly capable on mathematical and coding tasks~\citep{DBLP:journals/corr/abs-2602-08676, DBLP:journals/corr/abs-2512-13586, DBLP:journals/corr/abs-2602-12586}; given this strength, extending diffusion models to formal proving is a natural fit.

\textbf{Formal theorem proving and autoformalisation.} Distinct from general informal reasoning, formal reasoning has emerged as a promising direction for grounding LLM reasoning within formal proofs~\citep{leang-etal-2025-theorem, DBLP:conf/emnlp/ZhangVF25, DBLP:conf/emnlp/ZhangVF25a}. Several works advance formal reasoning~\citep{DBLP:journals/corr/abs-2405-14333, DBLP:journals/corr/abs-2507-08649, DBLP:journals/corr/abs-2504-06122, DBLP:journals/corr/abs-2502-07640, DBLP:conf/acl/XinXYCWXSZD25} through methods such as Monte Carlo Tree Search or breadth-first search, exploring multiple proof trajectories and iteratively assembling valid proofs to improve correctness. There is also recent work that applies formal reasoning beyond mathematics~\citep{DBLP:journals/corr/abs-2604-23002, DBLP:journals/corr/abs-2601-18944}. Recently, Seed-Prover~\citep{chen2025seed} uses techniques such as extensive test-time search and refinement to attain IMO medal-level performance; however, such systems rely on either large-scale training over very large base models or substantially higher computational overhead~\citep{chen2025seed}. Other lines of work explore bootstrapping~\citep{DBLP:conf/iclr/LinSWY25} and self-training~\citep{DBLP:conf/icml/Dong025a}. AlphaProof~\citep{1062014} introduces a mutation-based augmentation strategy to mitigate data scarcity in formal theorem proving, but the approach is compute-intensive: each candidate mutation requires Gemini~\citep{team2023gemini} to generate an accompanying proof and the Lean kernel to cross-check it before the variant can enter the training corpus. ALF removes this per-mutation overhead by operating at the statement level alone, retaining mutations on the basis of structural validity rather than verified proofs.
\method aims to mitigate these issues by developing small models without large computational cost, while also addressing the data scarcity that currently bottlenecks formal proving.

\section{Limitations and Future Work}
While our work makes several contributions, a number of natural extensions remain. (1) \emph{Diffusion-based formal proving.} Although \method-Diffusion currently trails its autoregressive counterpart, we believe diffusion models represent a promising research direction due to the mechanism of iterative refinement, and a meaningful architectural shift for LLM reasoning. Scaling the diffusion regime to longer effective contexts, and hybridising parallel diffusion refinement with autoregressive verification, are natural next steps. (2) \emph{Richer mutation operators.} Our data pipeline uses five mutation operators, which already yields a substantial diagnostic uplift on MiniF2F-ALF. We believe mutation is an efficient route to generating synthetic formal data, and that a broader operator family, including type-driven, lemma-graph, and proof-trace mutations, could further widen the failure surface that future provers must handle. (3) \emph{Beyond MiniF2F.} Our analysis indicates that MiniF2F is approaching saturation under contemporary Lean provers. Existing alternative benchmarks offer only partial substitutes: many are calibrated to a difficulty regime in which strong open-source language models register pass rates in the low single digits, while others have already approached saturation under current systems --- in either case providing limited resolution for distinguishing among the strongest provers. We position MiniF2F-ALF as a stronger drop-in variant within the MiniF2F family, but a broader community effort is needed to construct evaluation suites with controlled difficulty stratification across theorem-proving regimes.

\section{Conclusions}

We introduced \method, a compute-efficient family of open-source Lean~4 theorem provers spanning two generation paradigms: autoregressive provers at 4B and 32B parameters, and \method-Diffusion-4B, the first proof-of-concept diffusion-based theorem-proving model for Lean. The results show that strong formal proving need not rely exclusively on frontier-scale models: \method-4B outperforms a 671B-parameter prover on MiniF2F-Test despite being roughly $167\times$ smaller, while \method-32B achieves state-of-the-art performance on MiniF2F-Test and the strongest open-source result on PutnamBench. \method-Diffusion-4B shows that Lean proof generation is possible under a diffusion decoding regime, opening a complementary direction in which theorem provers can be compared not only by accuracy, but also by throughput and proofs per GPU-second.

The main driver of these results is a staged compute-frugal Lean data pipeline. Starting from a Lean-verified corpus we extend it with additional verified instances, and stratify it into easy, medium, and hard problems, which we use to train the models with a curriculum that exposes models progressively to more difficult proof tasks. We then introduce \emph{Augmented Lean Formalisation} (ALF), a structured mutation scheme that expands the verified corpus into diverse formal variants using lightweight sanity checks rather than per-instance Lean verification. The resulting variants serve both as self-distillation data for training and as the basis for \emph{MiniF2F-ALF}, a perturbation-based companion benchmark for evaluating robustness beyond exact benchmark memorisation. The degradation of contemporary provers on MiniF2F-ALF suggests that high accuracy on MiniF2F-Test does not yet imply robust transfer to structured formal variants. Together with parameter-efficient supervised fine-tuning, dynamic proof-reasoning filtering under an 8K context budget, self-distillation, and reinforcement learning from Lean verification, this data pipeline yields strong performance under practical compute budgets.

Overall, our results support the central premise of the paper: careful data construction and efficient training can substitute, in part, for raw model scale in Lean theorem proving.

\section*{Acknowledgements}
This work was supported by the AI for Math Fund, which is
managed by Renaissance Philanthropy in partnership with
founding donor XTX Markets. We are grateful to Huajian Xin, Chengqian Gao and Kwan Wai-Chung for helpful discussions on the training pipeline, and to Antonio Miceli Barone for valuable feedback on the manuscript.

\bibliographystyle{unsrtnat}
\bibliography{references}

\clearpage

\appendix

\setminted{
  fontsize=\tiny,
  breaklines=true,
  breakanywhere=true,
  breaksymbolleft={},
  breakautoindent=false,
  tabsize=2,
}

\section{Additional Details on Synthetic Data Generation}

\subsection{Rubric Distillation Prompt}
\label{app:analysis_lean}
For our rubric-based distillation stage, we use the prompt shown below to convert each failed Lean proof attempt into a self-contained synthetic mathematics problem with a step-by-step solution. The prompt instructs the generator to identify the root-cause error, classify it under our extraction rubric, and emit a strictly formatted JSON object containing the synthetic question and its solution.

\begin{tcolorbox}[
  breakable, enhanced,
  colback=gray!5, colframe=black!55,
  arc=2pt, boxrule=0.5pt,
  left=4pt, right=4pt, top=4pt, bottom=4pt,
  title={Prompt: ALF Failure-to-Synthetic-Problem Generation},
  fonttitle=\bfseries\small,
  fontupper=\footnotesize,
]
\begin{verbatim}
You are an expert mathematician and dataset creator. Your task is to
analyze a failed Lean 4 proof attempt, extract the exact mathematical
context where it failed, generate a WHOLE NEW SIMPLER SYNTHETIC, FULLY
SOLVABLE MATH PROBLEM, and provide its complete step-by-step
mathematical solution.

### CORE OBJECTIVE
Analyze a failed Lean 4 proof to identify the exact mathematical step
where the logic or formalization breaks down. Extract the specific
mathematical parameters, formulas, and premises from that step and the
original theorem statement. Use these extracted details to generate a
SIMPLER, SYNTHETIC, FULLY SOLVABLE MATH PROBLEM in pure natural
language, along with its step-by-step mathematical solution. This must
be a standalone mathematical question that contains all the necessary
numbers and context required to calculate the answer from scratch,
without needing to reference the original Lean code.

### ROOT CAUSE
When multiple errors are present, identify all the error by line
position as the root cause. However, ignore downstream errors that
would resolve once the root cause is fixed.
**Determine Failure Level:**
   - **Formalization error:** Malformed theorem statement (wrong
     types/fields). Target the underlying mathematical
     conceptualization of the claim.
   - **Proof error:** Bug inside the tactic block. Target mathematical
     strategy or logic.

### EXTRACTION RUBRIC (Map Error Category to Conceptual Math)
* **Invalid projection/field:** The proof calls a field that doesn't
  exist for that mathematical object. Extract the actual mathematical
  object and generate a problem that requires calculating or properly
  interacting with its true mathematical properties.
* **Unsolved goals:** Extract the exact remaining goal state and the
  available hypotheses. Turn this exact state into a self-contained
  "Prove that..." or "Calculate..." math problem.
* **Tactic failures (linarith, simp, rewrite, etc.):** Extract the
  specific algebraic equation or inequality that failed to simplify.
  Create a synthetic math problem asking the student to manually
  perform that specific algebraic manipulation or simplification.
* **Type mismatch:** Identify expected vs. actual domains (e.g.,
  integers vs. natural numbers, sets vs. finite collections). Create a
  problem that requires converting or mapping between these specific
  domains using the numbers from the theorem.
* **Failed to synthesize:** Missing mathematical property (e.g.,
  finiteness, decidability). Create a problem asking to prove that the
  specific object in the theorem possesses this required property.
* **Unknown constant/identifier:** The LLM hallucinated a theorem.
  Extract the logical step it was trying to make, and create a math
  problem that requires proving that specific logical step from first
  principles.
* **Other errors:** Extract the technical failure and translate it
  into a logic or calculation problem based on the active variables.

### CONSTRAINTS FOR THE SYNTHETIC QUESTION & SOLUTION
1. **COMPLETELY SELF-CONTAINED & SOLVABLE:** The generated question
   MUST include the actual mathematical values, coordinates, formulas,
   or premises involved in the failed step. (e.g., If the error
   involves Point F, you MUST state the exact coordinates of Point F
   in the question). It must be solvable without looking at the Lean
   code.
2. **PURE MATHEMATICS (NO CODE):** The question and solution MUST NOT
   contain Lean 4 code, tactics (`simp`, `rw`), Lean keywords (`let`,
   `have`), or backticks (` `). Frame it purely as a standard math
   competition or textbook problem.
3. **NO DIAGNOSTIC OR "WHY" QUESTIONS:** Do not ask "Why did this
   fail?" or "What is the correct logical approach?". Ask direct
   mathematical questions like "Calculate X", "Find Y", or "Prove that
   Z is true given [Premises]."
4. **EXPLICIT STEP-BY-STEP SOLUTION:** You must provide a clear,
   logically sound, step-by-step mathematical solution to the
   synthetic question you just generated.

### EXAMPLES (Bad vs. Good)
- **BAD (Diagnostic/Code-heavy):** "Why does the `intro` tactic fail
  when applied to Point F?"
- **GOOD (Synthetic Math Problem):** "Point A is at (0, 0) and Point B
  is at (8/3, 4). If Point F is defined as the midpoint of segment AB,
  calculate the exact sum of the x and y coordinates of Point F."

- **BAD (Diagnostic/Vague):** "What missing logical step is required
  to prove that this sum is strictly positive?"
- **GOOD (Synthetic Math Problem):** "Given that x > 5 and y = -2,
  prove that the sum of x and y is strictly positive."

---
### BEGIN USER PROMPT
PREVIOUS QUERY
{{QUESTION}}

FORMAL STATEMENT:
```lean4
{{FORMAL_STATEMENT}}

FAILED PROOF:


{{MERGED_PROOF}}
ERROR MESSAGE:
{{ERROR_MESSAGE}}
\end{verbatim}
\end{tcolorbox}

\subsection{Similarity Filtering for Intra-corpus Deduplication}
\label{app:similarity}

For the \emph{easy} tier we deduplicate by semantic content rather than surface form, since paraphrase-level duplicates are common in our problem sources and surface methods (exact-match or normalised edit distance) miss variable-renamed or lightly reworded copies. Each statement is encoded with \texttt{Alibaba-NLP/gte-Qwen2-7B-instruct}~\citep{li2023towards} into a $3{,}584$-dimensional dense embedding. Embeddings are $\ell_2$-normalised so that cosine similarity reduces to an inner product, and indexed with FAISS~\citep{DBLP:journals/tbd/DouzeGDJSMLHJ26} using an inner-product index for efficient $k$-nearest-neighbour search.

\textbf{Chunked overlapping search.} Exhaustive pairwise comparison across the full easy-tier pool is memory-bound, so we partition the embedded instances into chunks of $5{,}000$ items with a $100$-instance overlap between consecutive chunks, and run top-$k$ nearest-neighbour retrieval within each chunk. The overlap guarantees that near-duplicates whose embeddings happen to straddle a chunk boundary are still compared in at least one chunk, preventing duplicates from leaking through partitioning.

\textbf{Filtering rule.} A pair $(x_i, x_j)$ with cosine similarity above $\tau$ is treated as a near-duplicate, and the lower-priority member is dropped by source-tier priority (curated problems over autoformalised paraphrases; older sources over newer reformulations). We set $\tau = 0.75$ via a small held-out sweep that balances false-positive removal (distinct problems flagged as duplicates) against false-negative retention (near-duplicates passing through).

\subsection{Dataset Decontamination Sweep}
\label{app:decontamination}

Beyond intra-corpus deduplication (Appendix~\ref{app:similarity}), we run a five-stage decontamination sweep against every benchmark we report on (MiniF2F-Valid, MiniF2F-Test, and PutnamBench); a training instance is discarded if any stage triggers a hit. (i) \emph{Surface-form hashing:} we normalise the natural-language statement (lowercase, collapsed whitespace, stripped punctuation, canonicalised numerals) and the Lean statement ($\alpha$-renamed binders, comments and formatting removed) and compare MD5 hashes. (ii) \emph{$n$-gram overlap:} we flag any instance sharing an $n$-gram with a benchmark item, with $n=10$ on natural language and $n=5$ on Lean (the shorter window reflects the lower lexical entropy of formal code). (iii) \emph{Embedding similarity:} surviving candidates are encoded with \texttt{gte-Qwen2-7B-instruct} and indexed in FAISS as in Appendix~\ref{app:similarity}, and removed if their cosine similarity to any benchmark statement exceeds $\tau_{\text{decon}}$. (iv) \emph{Structural equivalence:} we compare $\alpha$-normalised Lean statement trees and remove any instance that elaborates to the same formal theorem as a benchmark item despite a different surface form. (v) \emph{Proof-template overlap:} as a cheaper alternative to retraining-based checks, we strip variable names and constants from both training and benchmark proofs, normalise the residual tactic skeletons, and flag any training instance whose skeleton is near-identical to that of a benchmark proof, catching cases where a benchmark proof is reproduced up to renaming.

Because MiniF2F-ALF is obtained by mutating MiniF2F-Test, any MiniF2F-Test contamination is also an ALF-seed contamination; we additionally run the first three stages against the $488$ mutated ALF statements to catch the residual case where a training instance matches a mutated variant rather than the original. The resulting decontaminated set is used for all supervised fine-tuning, self-distillation, and reinforcement-learning runs reported in this paper.

\subsection{ALF Self-Distillation}
For our ALF Self-Distillation generation, each mutated statement's formal proof is generated by \method-Post-RL teacher at temperature $T=1.2$ and nucleus $\text{top-}p=0.95$, and the completion is used as a training target. Given the scale (2M instances per model), we omit per-sample Lean verification; \citet{DBLP:journals/corr/abs-2604-01193} report that this omission does not materially degrade downstream performance in the coding domain, and Appendix~\ref{app:decomposition} indicates the same holds for formal proving, with self-distillation contributing consistently on top of the SFT baseline at both 4B and 32B.

\subsection{Verification of ALF Synthetic Data}
Verifying all $2$M ALF self-distilled instances with the Lean compiler is computationally prohibitive. We therefore conduct a random audit: sampling $2{,}000$ retained ALF instances uniformly from the final self-distillation corpus and checking each complete generated theorem, including both the mutated formal statement and its proof, under the same Lean environment used for evaluation. Of these audited instances, $87.8\%$ compile successfully, indicating that the quality of the synthetic data remains high and that exhaustive verification is unnecessary for our purposes. We do, however, verify the formal \emph{statement} alongside the formal proof rather than the proof alone: checking only the proof teaches the model to alter the statement to fit a flawed proof, an effect we observe in practice and find necessary to prevent. Full verification of every instance is left out only on grounds of compute.

\subsubsection{Why Self-Distillation?}

After supervised fine-tuning, \method\ already attains a strong pass rate on MiniF2F-Test (Table~\ref{tab:pythagoras_decomposition}); the remaining headroom is not a question of base capability but of exposing the model to a wider distribution of proof trajectories. Preference-based post-training is poorly matched to this regime: Direct Preference Optimisation~\citep{DBLP:conf/nips/RafailovSMMEF23} expects a dense preference signal, but the Lean verifier returns only a sparse binary judgement, and the preference pairs we derived from it proved unstable in our runs. GRPO suffers the same scarcity, yielding only marginal gains because the binary success/fail reward leaves little gradient on the long tail of structurally distinct but equally valid proofs. Self-distillation via ALF sidesteps both issues. It supplies dense, full-trajectory supervision: the teacher emits a complete proof for each mutated statement, rather than a single bit of reward. It \emph{expands} rather than refines the training distribution, drawing formally distinct theorem variants from each seed. 
Self-distillation is therefore the right tool for this regime: full proof trajectories are more informative than sparse preference signals, and ALF is the mechanism that turns one seed into many.
\subsection{Examples of \method-Dataset for Each Difficulty}
\label{sec:difficulty}

\begin{tcolorbox}[
  enhanced, sidebyside, sidebyside align=top, sidebyside gap=4mm,
  colback=green!4, colframe=green!50!black,
  title={\textbf{Easy tier example.} Synthetic, rubric-simplified algebra problem (rectangle perimeter).},
  fonttitle=\small,
  left=3pt, right=3pt, top=3pt, bottom=3pt,
]
{\footnotesize\textbf{Formal statement}}
\begin{minted}{lean4}
import Mathlib
import Aesop

set_option maxHeartbeats 0

open BigOperators Real Nat Topology Rat

/-- Let `a` and `b` be positive real numbers
    such that `b = a + 9` and `a * b = 360`.
    Prove that `2 * (a + b) = 78`. -/
theorem my_algebra_theorem
    (a b : ℝ) (ha : a > 0) (hb : b > 0)
    (h1 : b = a + 9) (h2 : a * b = 360) :
    2 * (a + b) = 78 := by
  sorry
\end{minted}
\tcblower
{\footnotesize\textbf{Formal proof}}
\begin{minted}{lean4}
theorem my_algebra_theorem
    (a b : ℝ) (ha : a > 0) (hb : b > 0)
    (h1 : b = a + 9) (h2 : a * b = 360) :
    2 * (a + b) = 78 := by
  have h3 : a * (a + 9) = 360 := by
    rw [← h1]; exact h2
  have ha15 : a = 15 := by
    nlinarith [sq_nonneg (a - 15), sq_nonneg (a + 24), ha]
  have hb24 : b = 24 := by
    rw [h1, ha15]; norm_num
  rw [ha15, hb24]; norm_num

\end{minted}
\end{tcolorbox}
\begin{tcolorbox}[
  enhanced, sidebyside, sidebyside align=top, sidebyside gap=4mm,
  colback=orange!4, colframe=orange!60!black,
  title={\textbf{Medium tier example.} Autoformalised general-reasoning word problem (camper marshmallow count).},
  fonttitle=\small,
  left=3pt, right=3pt, top=3pt, bottom=3pt,
]
\begin{minted}{lean4}
import Mathlib
import Aesop

set_option maxHeartbeats 0

open BigOperators Real Nat Topology Rat

/-- At camp Wonka there are 96 campers.
    Two-thirds are boys, one-third are girls.
    50%
    a marshmallow. How many marshmallows are
    needed in total? -/
theorem my_combinatorics_theorem :
    let total_campers       : ℕ := 96
    let boys                : ℕ := (2 * total_campers) / 3
    let girls               : ℕ := total_campers / 3
    let boys_wanting_toast  : ℕ := boys / 2
    let girls_wanting_toast : ℕ := (3 * girls) / 4
    let total_marshmallows  : ℕ :=
      boys_wanting_toast + girls_wanting_toast
    total_marshmallows = 56 := by
  sorry
\end{minted}
\tcblower
\begin{minted}{lean4}
theorem my_combinatorics_theorem :
    let total_campers       : ℕ := 96
    let boys                : ℕ := (2 * total_campers) / 3
    let girls               : ℕ := total_campers / 3
    let boys_wanting_toast  : ℕ := boys / 2
    let girls_wanting_toast : ℕ := (3 * girls) / 4
    let total_marshmallows  : ℕ :=
      boys_wanting_toast + girls_wanting_toast
    total_marshmallows = 56 := by
  intro total_campers boys girls
        boys_wanting_toast girls_wanting_toast
        total_marshmallows
  have h_boys : boys = 64 := by
    dsimp [boys, total_campers] <;> norm_num <;> rfl
  have h_girls : girls = 32 := by
    dsimp [girls, total_campers] <;> norm_num <;> rfl
  have h_boys_toast : boys_wanting_toast = 32 := by
    dsimp [boys_wanting_toast, boys, total_campers] at *
    <;> norm_num at * <;> rfl
  have h_girls_toast : girls_wanting_toast = 24 := by
    dsimp [girls_wanting_toast, girls, total_campers] at *
    <;> norm_num at * <;> rfl
  have h_total : total_marshmallows = 56 := by
    dsimp [total_marshmallows,
           boys_wanting_toast, girls_wanting_toast] at *
    <;> norm_num [h_boys_toast, h_girls_toast] at *
    <;> rfl
  exact h_total
\end{minted}
\end{tcolorbox}

\begin{tcolorbox}[
  enhanced, sidebyside, sidebyside align=top, sidebyside gap=4mm,
  colback=red!4, colframe=red!55!black,
  title={\textbf{Hard tier example.} Competition-style number theory: smallest prime $p$ such that $p^{2}+144$ is a perfect square.},
  fonttitle=\small,
  left=3pt, right=3pt, top=3pt, bottom=3pt,
]
\begin{minted}{lean4}
import Mathlib
import Aesop

set_option maxHeartbeats 0

open BigOperators Real Nat Topology Rat

/-- Let `p` be the smallest prime such that
    `p^2 + 144` is also a perfect square.
    Determine the value of `p`. -/
theorem my_favorite_prime :
    Nat.Prime 5 ∧
    (∃ k : ℕ, 5^2 + 144 = k^2) ∧
    (∀ p : ℕ, Nat.Prime p → p < 5 →
      ¬∃ k : ℕ, p^2 + 144 = k^2) := by
  sorry
\end{minted}

\tcblower
\begin{minted}{lean4}
theorem my_favorite_prime :
    Nat.Prime 5 ∧
    (∃ k : ℕ, 5^2 + 144 = k^2) ∧
    (∀ p : ℕ, Nat.Prime p → p < 5 →
      ¬∃ k : ℕ, p^2 + 144 = k^2) := by
  have h_prime_5 : Nat.Prime 5 := by decide
  have h_exists_k : ∃ k : ℕ, 5^2 + 144 = k^2 := by
    refine' ⟨13, _⟩
    norm_num
  have h_no_smaller_prime :
      ∀ p : ℕ, Nat.Prime p → p < 5 →
        ¬∃ k : ℕ, p^2 + 144 = k^2 := by
    intro p hp hlt
    have h₁ : p = 2 ∨ p = 3 := by
      have h₄ : p ≤ 4 := by linarith
      interval_cases p <;>
        norm_num [Nat.Prime] at hp ⊢ <;> aesop
    rcases h₁ with (rfl | rfl)
    · intro h
      rcases h with ⟨k, hk⟩
      have h₂ : 2 ^ 2 + 144 = k ^ 2 := by simpa using hk
      have h₃ : k ^ 2 = 148 := by linarith
      have h₄ : k ≤ 12 := by nlinarith
      have h₅ : k ≥ 13 := by nlinarith
      linarith
    · intro h
      rcases h with ⟨k, hk⟩
      have h₂ : 3 ^ 2 + 144 = k ^ 2 := by simpa using hk
      have h₃ : k ^ 2 = 153 := by linarith
      have h₄ : k ≤ 12 := by nlinarith
      have h₅ : k ≥ 13 := by nlinarith
      linarith
  exact ⟨h_prime_5, h_exists_k, h_no_smaller_prime⟩
\end{minted}

\end{tcolorbox}

\section{Experimental Settings}
\label{app:settings}

\subsection{Training Details}
\label{app:training-details}
All models are based on the Qwen3 series~\citep{DBLP:journals/corr/abs-2505-09388}, specifically Qwen3-4B and Qwen3-32B, and are fine-tuned for two epochs using LoRA of rank $64$ applied to all linear layers. Our models are trained using LlamaFactory~\citep{zheng2024llamafactory} for SFT, and Volcano Engine Reinforcement Learning (verl)~\citep{sheng2024hybridflow} for RL. We use learning rates of $3\mathrm{e}{-4}$ and $7\mathrm{e}{-5}$ for the SFT stages. For self-distillation, we reduce the learning rate to $1\mathrm{e}{-5}$, as larger values induced training instability. Lastly, for \method-Diffusion we perform full fine-tuning with a learning rate of $3\mathrm{e}{-5}$.

\subsection{MiniF2F-ALF Creation}
\label{app:minif2f_settings}
MiniF2F-ALF is constructed by applying the ALF mutation scheme of \S\ref{sec:alf} to each of the $244$ problems in MiniF2F-Test. The design follows GSM-Symbolic~\citep{DBLP:conf/iclr/MirzadehASTBF25}, which shows that perturbing a problem's variables and numerical values while preserving its underlying structure and difficulty is sufficient to expose memorisation effects on GSM8K; we transfer this principle to formal proving by renaming variables in each statement, reducing reliance on memorised surface forms. For each problem we generate five candidate mutations with Codex (GPT-5.5) and rank them by semantic divergence from the original statement, measured as cosine distance under \texttt{Alibaba-NLP/gte-Qwen2-7B-instruct}~\citep{li2023towards} (the encoder of Appendix~\ref{app:similarity}), retaining the two most divergent candidates per problem. Unlike the ALF self-distillation corpus, where per-sample verification is omitted at scale (\S\ref{sec:alf}), here every retained mutation is verified: we use Claude Opus 4.6~\citep{anthropic2026claude46systemcard} as a judge to confirm that it is a well-formed Lean theorem and pass it through the Lean compiler to ensure correctness, so that all $488$ statements in the final benchmark are guaranteed valid.

\begin{table}[t]
\centering
\caption{Pass@$N$ on MiniF2F-Test across inference budgets~$N$. Results in the top part are copied as reported by the respective papers. Values are percentages; $^\dagger$ Kimina-Prover variants; best in bold.}
\label{tab:budget_performance_full}
\small
\setlength{\tabcolsep}{8pt}
\renewcommand{\arraystretch}{1.12}
\resizebox{0.75\linewidth}{!}{%
\begin{tabular}{@{}p{0.45\linewidth} >{\raggedleft\arraybackslash}p{0.13\linewidth} >{\raggedleft\arraybackslash}p{0.24\linewidth}@{}}
\toprule
\textbf{Method} & \textbf{Budget ($N$)} & \textbf{Performance Pass@$N$} \\
\midrule
Goedel-Prover-SFT~\citep{DBLP:journals/corr/abs-2502-07640}
    & 32   & 57.6 \\
    & 3200 & 62.7 \\
\midrule
STP~\citep{DBLP:conf/icml/Dong025a}
    & 128   & 61.2 \\
    & 3200  & 65.0 \\
    & 25600 & 67.6 \\
\midrule
Kimina-Prover-Preview-72B~\citep{wang2025kiminaproverpreviewlargeformal}
    & 32   & 68.85 \\
    & 8192 & 80.74 \\
\midrule
DeepSeek-Prover-V2-7B~\citep{DBLP:journals/corr/abs-2504-21801}
    & 32   & 75.6 \\
    & 8192 & 82.0 \\
\cdashline{1-3}\\[-0.6em]
DeepSeek-Prover-V2-671B
    & 32   & 82.4 \\
    & 8192 & 88.9 \\
\midrule
Kimina-Prover-8B$^\dagger$~\citep{wang2025kiminaproverpreviewlargeformal}
    & 32 & 78.3 \\
\cdashline{1-3}\\[-0.6em]
Kimina-Prover-70B$^\dagger$
    & 32   & 84.0 \\
    & 1024 & 87.7 \\
\cdashline{1-3}\\[-0.6em]
\hspace{1em}w/ Test-Time Reinforcement Learning
    & --   & 92.2 \\
\midrule
Goedel-Prover-V2-8B~\citep{lin2025goedelproverv2scalingformaltheorem}
    & 32   & 84.6 \\
    & 1024 & 87.9 \\
    & 8192 & 90.2 \\
\cdashline{1-3}\\[-0.6em]
\hspace{1em}w/ Self-Correction
    & 32   & 86.7 \\
    & 1024 & 89.3 \\
\cdashline{1-3}\\[-0.6em]
Goedel-Prover-V2-32B
    & 32   & 88.1 \\
    & 1024 & 91.8 \\
    & 8192 & 92.2 \\
\cdashline{1-3}\\[-0.6em]
\hspace{1em}w/ Self-Correction
    & 32   & 90.4 \\
    & 1024 & 92.6 \\
\midrule
\rowcolor{gray!15}
\textbf{\method-4B}
    & 32   & 86.1 \\
\rowcolor{gray!15}
    & 1024 & 88.1 \\
\rowcolor{gray!15}
    & 2048 & 89.8 \\
\midrule
\rowcolor{gray!15}
\textbf{\method-32B}
    & 32   & 89.8 \\
\rowcolor{gray!15}
    & 1024 & 92.6 \\
\rowcolor{gray!15}
    & 2048 & \textbf{93.0} \\
\bottomrule
\end{tabular}%
}
\end{table}
\section{Extended Results and Ablations}
\subsection{Full Pass@N Results}

Table~\ref{tab:budget_performance_full} reports pass@$N$ on MiniF2F-Test across the full sweep of inference budgets we evaluate, covering every open-source baseline both with and without self-correction.

At the 4B scale, \method-4B reaches $86.1\%$ at pass@$32$. This places it ahead of Goedel-Prover-V2-8B without self-correction ($84.6\%$) and at near parity with that model's self-corrected variant ($86.7\%$), despite using half the parameters and no inference-time correction loop. The advantage widens with budget: \method-4B attains $88.1\%$ at pass@$1024$ and $89.8\%$ at pass@$2048$, the latter already exceeding the pass@$8192$ result of DeepSeek-Prover-V2-671B ($88.9\%$) at a quarter of the sampling budget and with roughly $167\times$ fewer parameters.

At the 32B scale, \method-32B leads Goedel-Prover-V2-32B without self-correction by $1.7$ points at pass@$32$ ($89.8\%$ vs.\ $88.1\%$), trails its self-corrected counterpart by only $0.6$ points at the same budget ($89.8\%$ vs.\ $90.4\%$), and matches it at pass@$1024$ ($92.6\%$ vs.\ $92.6\%$). At pass@$2048$ it reaches $\mathbf{93.03\%}$, the strongest reported pass rate on MiniF2F-Test, surpassing the pass@$8192$ result of Goedel-Prover-V2-32B ($92.2\%$) at a quarter of the budget and without any iterative repair step.

\subsection{Decomposition Performance}
\label{app:decomposition}

\begin{table}[ht]
\centering
\caption{Ablation of \method training components on MiniF2F-Test under a pass@32 inference budget. SFT-only rows isolate the contribution of the synthetic-data supervised fine-tuning stage prior to ALF self-distillation.}
\label{tab:pythagoras_decomposition}
\small
\setlength{\tabcolsep}{8pt}
\renewcommand{\arraystretch}{1.12}
\begin{tabular}{p{0.56\linewidth}cc}
\toprule
\textbf{Method} & \textbf{Budget} & \textbf{Pass@32 (\%)} \\
\midrule

\rowcolor{gray!15}
\method-4B (SFT only)
    & 32 & $79.10$ \\
\cdashline{1-3}

\rowcolor{gray!15}
\method-4B (SFT + ALF Self-Distillation)
    & 32 & $86.07$ \\
\midrule

\rowcolor{gray!15}
\method-32B (SFT only)
    & 32 & $84.02$ \\
\cdashline{1-3}

\rowcolor{gray!15}
\method-32B (SFT + ALF Self-Distillation)
    & 32 & $89.75$ \\
\bottomrule
\end{tabular}
\end{table}

Table~\ref{tab:pythagoras_decomposition} decomposes the contributions of supervised fine-tuning and ALF self-distillation on MiniF2F-Test at pass@32. Supervised fine-tuning on our synthetic-data corpus alone already yields a strong prover ($79.10\%$ at 4B and $84.02\%$ at 32B) which we read as direct evidence that the corpus itself carries most of the signal required for competitive formal reasoning, prior to any further training stages. Adding ALF self-distillation lifts the two models to $86.1\%$ and $89.8\%$, gains of $+6.97$ and $+5.73$ points respectively, showing that mutation-driven self-distillation delivers a consistent improvement on top of an already strong SFT baseline rather than substituting for it.

\subsection{Why Long-Context Diffusion Training Destabilises}
\label{app:long_context_instability}
\begin{figure}[ht]
  \centering
  \resizebox{0.85\textwidth}{!}{%
    \begin{minipage}{\textwidth}
      \begin{subfigure}[t]{0.48\linewidth}
        \centering
        \includegraphics[width=\linewidth]{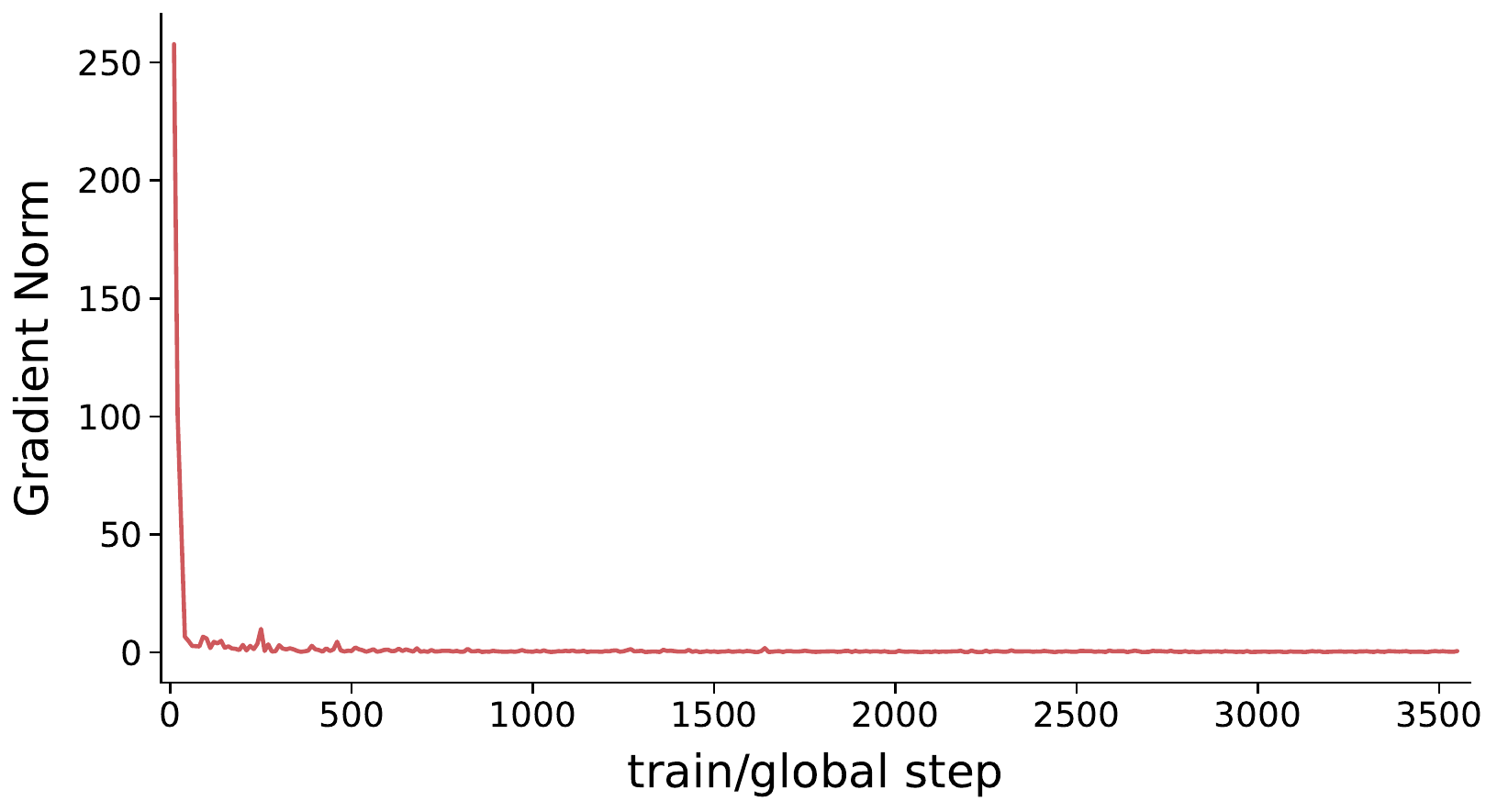}
        \caption{Stable training at $4{,}096$-token context: gradient norm settles after warm-up and remains bounded throughout training.}
        \label{fig:diff_gradnorm_stable}
      \end{subfigure}\hfill
      \begin{subfigure}[t]{0.48\linewidth}
        \centering
        \includegraphics[width=\linewidth]{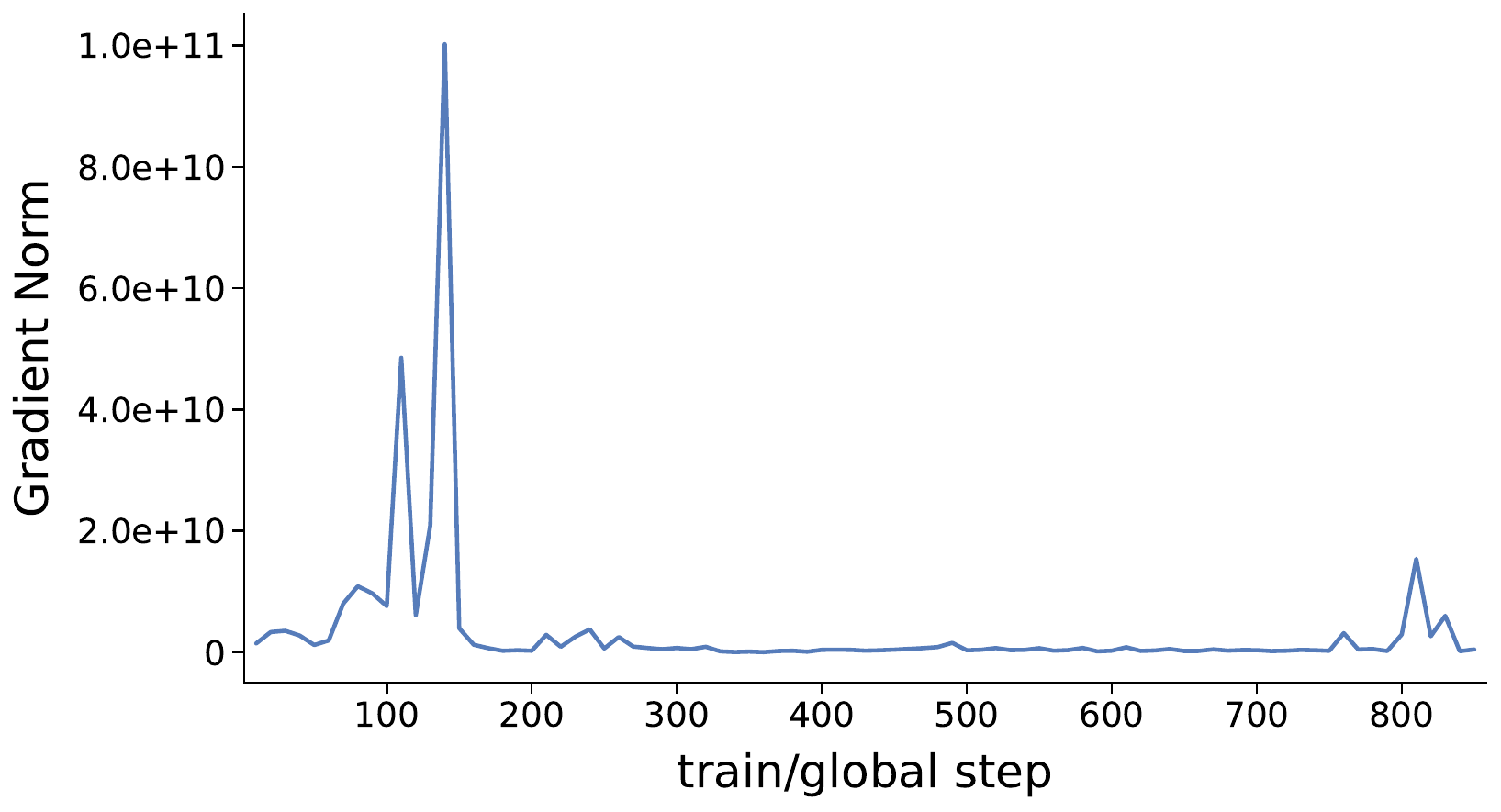}
        \caption{Unstable training at $8{,}192$-token context: gradient norm fails to settle and exhibits sustained spikes well past warm-up.}
        \label{fig:diff_gradnorm_unstable}
      \end{subfigure}
    \end{minipage}%
  }
  \caption{Gradient-norm dynamics of \method-Diffusion-4B under the two context-length settings. The $8{,}192$-token setting that converges cleanly under the autoregressive objective destabilises diffusion training, motivating the reduced $4{,}096$-token budget used for \method-Diffusion (\S\ref{sec:scaling_diffusion}).}
  \label{fig:diff_gradnorm}
\end{figure}
The instability illustrated in Figure~\ref{fig:diff_gradnorm} is not a tuning artefact but a consequence of the variance structure of the MDLM objective. We give the analysis here, and connect it to the design choices reported in prior masked-diffusion LM work.

\paragraph{Variance of the per-sequence loss.}
For a fixed sequence $\mathbf{x}_0$ of length $L$ and fixed noise level $t$, the forward process masks each position independently with probability $t$, so the number of masked positions is $|S| \sim \mathrm{Binomial}(L, t)$. The MDLM stochastic loss for this sample is
\begin{equation}
    \widehat{\mathcal{L}}(\theta; \mathbf{x}_0, \mathbf{x}_t, t) \;=\; -\frac{1}{t}\sum_{i=1}^{L} \mathbf{1}\!\left[x_t^i = [\textsc{m}]\right] \log p_\theta(x_0^i \mid \mathbf{x}_t).
    \label{eq:mdlm-stochastic}
\end{equation}
Treating the per-position log-probabilities as bounded by an average magnitude $\bar\ell$, the conditional variance over the masking draw is
\begin{equation}
    \mathrm{Var}\!\left[\widehat{\mathcal{L}} \,\middle|\, t, \mathbf{x}_0\right] \;\approx\; \bar\ell^{\,2} \cdot \frac{L\,(1-t)}{t}.
    \label{eq:mdlm-variance}
\end{equation}
Here $\bar\ell$ denotes a representative per-position log-probability magnitude, $\bar\ell \approx \lvert\log p_\theta(x_0^i \mid \mathbf{x}_t)\rvert$, treated as roughly constant across positions. We stress that Eq.~\eqref{eq:mdlm-variance} is approximate: collapsing the per-position losses to a single scale $\bar\ell$ treats them as independent and omits their cross-position covariance, whereas in practice the $\log p_\theta(x_0^i \mid \mathbf{x}_t)$ are correlated, since all masked positions are predicted from the same partially-masked context. We therefore read Eq.~\eqref{eq:mdlm-variance} as a behavioural scaling, $\mathrm{Var}\propto L(1-t)/t$, rather than an exact variance.

Although Eq.~\eqref{eq:mdlm-variance} is the variance of the loss, the gradient $\nabla_\theta\widehat{\mathcal{L}}$ shares the same $1/t$-reweighted sum-over-masked-positions structure, so its variance carries the same $L(1-t)/t$ scaling; this is the quantity reflected in the gradient-norm spikes of Figure~\ref{fig:diff_gradnorm_unstable}.

Two properties of Eq.~\ref{eq:mdlm-variance} matter for stability: (i) the variance is \emph{linear in the sequence length $L$}, so doubling the context from $4{,}096$ to $8{,}192$ doubles per-step gradient variance at every noise level; and (ii) the variance \emph{diverges as $t \to 0$}, so rare low-$t$ realisations — which are unavoidable under the uniform schedule $t \sim \mathcal{U}[0,1]$ used by LLaDA~\citep{DBLP:journals/corr/abs-2502-09992} and by Diffusion-\method — produce heavy-tailed contributions whose magnitude also scales with $L$. The compound effect is that long-context training amplifies both the typical gradient variance and the tail mass of large-gradient steps, so even short bursts of low-$t$ samples are enough to push the optimiser past gradient clipping in sustained spikes, which is precisely the pattern visible in Figure~\ref{fig:diff_gradnorm_unstable}.

This effect compounds with the memory budget: doubling the context roughly halves the largest feasible batch size, so the per-sequence variance increase of Eq.~\eqref{eq:mdlm-variance} is averaged over fewer samples per step, further raising the gradient variance the optimiser actually sees.

\paragraph{Extension to tactic-based masking.}
The analysis above is stated for the token-level objective, 
whereas \method-Diffusion masks whole tactic spans (Eq.~\eqref{eq:tac-loss}). The same argument transfers with the unit of masking changed from tokens to tactics. Writing $K$ for the number of tactic spans and $\ell_{\tau_k}\approx|\tau_k|\,\bar\ell$ for the summed per-token loss of span $\tau_k$, the per-sequence variance becomes
\begin{equation}
\mathrm{Var}\!\left[\widehat{\mathcal{L}}_{\mathrm{tac}}\,\middle|\,t, \mathbf{x}_0\right] \;\approx\; \frac{1-t}{t}\sum_{k=1}^{K}\ell_{\tau_k}^{2} \;\approx\; \bar\ell^{\,2}\,\bar{s}\,\frac{L(1-t)}{t},
\end{equation}
where $\bar{s}$ is the mean tactic length (with equality when spans are equal-length; variable span lengths increase the bound further). Tactic-based masking therefore inherits the same $L$- and $t$-dependence and amplifies it by roughly $\bar{s}$, so the instability is, if anything, more pronounced for \method-Diffusion than for a token-level masked-diffusion model at the same context length.

\paragraph{Contrast with autoregressive training.}
This variance is specific to the diffusion objective. The autoregressive loss $-\sum_{i}\log p_\theta(x_0^i\mid \mathbf{x}_0^{<i})$ is a deterministic sum over all $L$ positions, with no masking draw and no $1/t$ reweighting, so it carries none of the $L(1-t)/t$ terms. This is why the same $8{,}192$-token setting that destabilises diffusion converges cleanly under the autoregressive objective (\S\ref{sec:scaling_diffusion}).

\paragraph{Consistency with prior masked-diffusion LM work.}
Every existing masked-diffusion LM that scales to large models trains at a relatively short, fixed context window and explicitly engineers around the variance term of Eq.~\ref{eq:mdlm-variance}. LLaDA~\citep{DBLP:journals/corr/abs-2502-09992} pretrains at a \emph{fixed} $4{,}096$-token context and reserves $1\%$ of the data for random shorter lengths, with no recipe given for $8{,}192$. MDLM~\citep{sahoo2024simple} reduces ELBO variance through a low-discrepancy $t$-sampler that draws correlated $t_i$ rather than i.i.d.. Block diffusion~\citep{DBLP:conf/iclr/ArriolaGCYQHSK25} goes further still, introducing gradient-variance estimators and data-driven noise schedules and decomposing the sequence into smaller blocks so that no single forward pass has to absorb the full $L(1-t)/t$ variance term. The shared pattern across these works is that long-sequence stability in masked diffusion is treated as a \emph{variance-reduction} problem rather than an optimiser-hyperparameter problem, which matches our empirical observation: the only intervention that consistently stabilised \method-Diffusion at $8{,}192$ tokens in our preliminary experiments was reducing the context budget, while increased warm-up, lower learning rate, and tighter gradient clipping each only postponed the spike pattern of Figure~\ref{fig:diff_gradnorm_unstable}. We therefore use the $4{,}096$-token budget throughout for \method-Diffusion, and leave variance-reduction techniques (low-discrepancy / antithetic $t$-sampling, block decomposition) to future work.

\subsection{Training Performance between full context and dynamic \emph{proof-reasoning filtering}}
\label{sec:dynamic_filtering_ablation}
In this section, we perform an ablation study between full-context training and our full-context dynamic training with proof-reasoning filtering. Both settings share the same model backbone, synthetic-data corpus, optimiser configuration, and SFT recipe; they differ only in whether reasoning traces are filtered on the fly during training. The full-context baseline trains on every reasoning--proof pair extracted from the synthetic corpus, whereas the dynamic-filtering variant prunes each minibatch in real time by discarding pairs whose reasoning trace is inconsistent with its closing tactic sequence, concentrating gradient updates on traces whose chain-of-thought actually supports the discharged theorem.

Table~\ref{tab:filtering_ablation} reports pass@32 on MiniF2F-Test under both settings. The dynamic filter delivers a consistent improvement at both scales, $+3$ solved instances at 4B and $+2$ instances at 32B, without changing parameter count, training budget, or inference protocol. The gain is meaningful in the near-saturated regime of MiniF2F-Test: the 4B model picks up three additional problems on a benchmark where the four-prover panel of \S\ref{sec:minif2f_analysis} leaves only $41/244$ at-least-one-wrong cases to contest. The smaller absolute gain at 32B is consistent with the diminishing-returns pattern observed elsewhere in the analysis: the 32B baseline already discharges most reasoning-noisy traces robustly, so filtering shifts mass on a narrower residual.

\begin{table}[ht]
\centering
\caption{Ablation of dynamic proof-reasoning filtering against full-context training on MiniF2F-Test at pass@32. Both rows share the SFT recipe and synthetic-data corpus; the only difference is whether reasoning traces are filtered on the fly during training. Values are pass rates (\%) with absolute solved counts out of 244 in parentheses.}
\label{tab:filtering_ablation}
\resizebox{0.8\textwidth}{!}{%
\begin{tabular}{lcc}
\toprule
\textbf{Training setting} & \textbf{\method-4B} & \textbf{\method-32B} \\
\midrule
Full context (no filtering)                & 80.70 & 84.02 \\
\;+ Dynamic proof-reasoning filtering      &
79.10 & 83.20 \\
\midrule
$\Delta$                                   & $+1.62$ & $+0.82$ \\
\bottomrule
\end{tabular}%
}
\end{table}

\section{Detailed Analysis of MiniF2F}
\label{app:further_analyse_minif2f}

\begin{figure}[ht]
    \centering
    \begin{subfigure}[t]{0.31\textwidth}
        \centering
        \includegraphics[width=\linewidth,height=0.23\textheight,keepaspectratio]{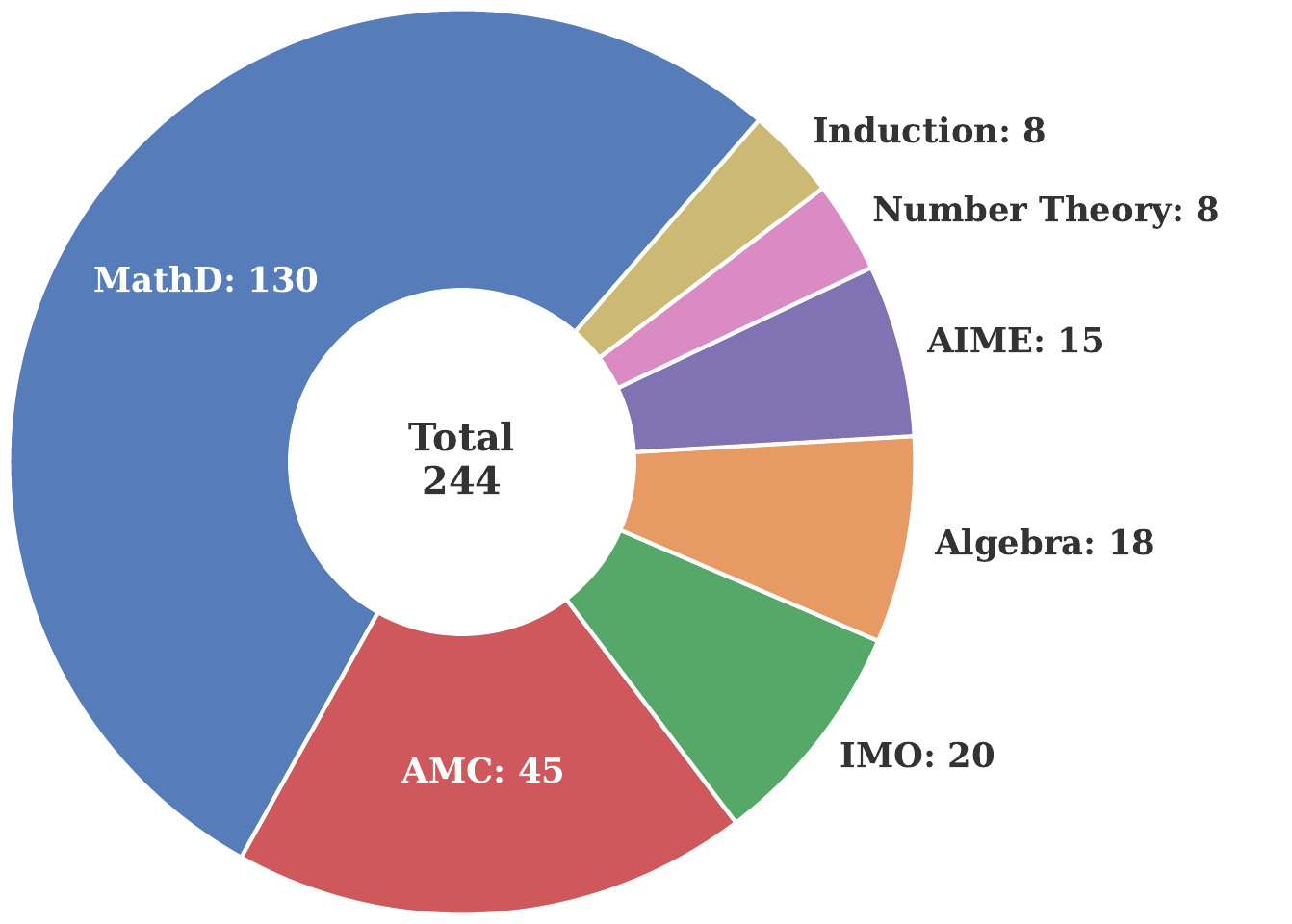}
        \caption{Domain composition of the full MiniF2F-Test benchmark (all instances).}
        \label{fig:minif2f_full_piechart}
    \end{subfigure}
    \hspace{0.02\textwidth}
    \begin{subfigure}[t]{0.31\textwidth}
        \centering
        \includegraphics[width=\linewidth,height=0.23\textheight,keepaspectratio]{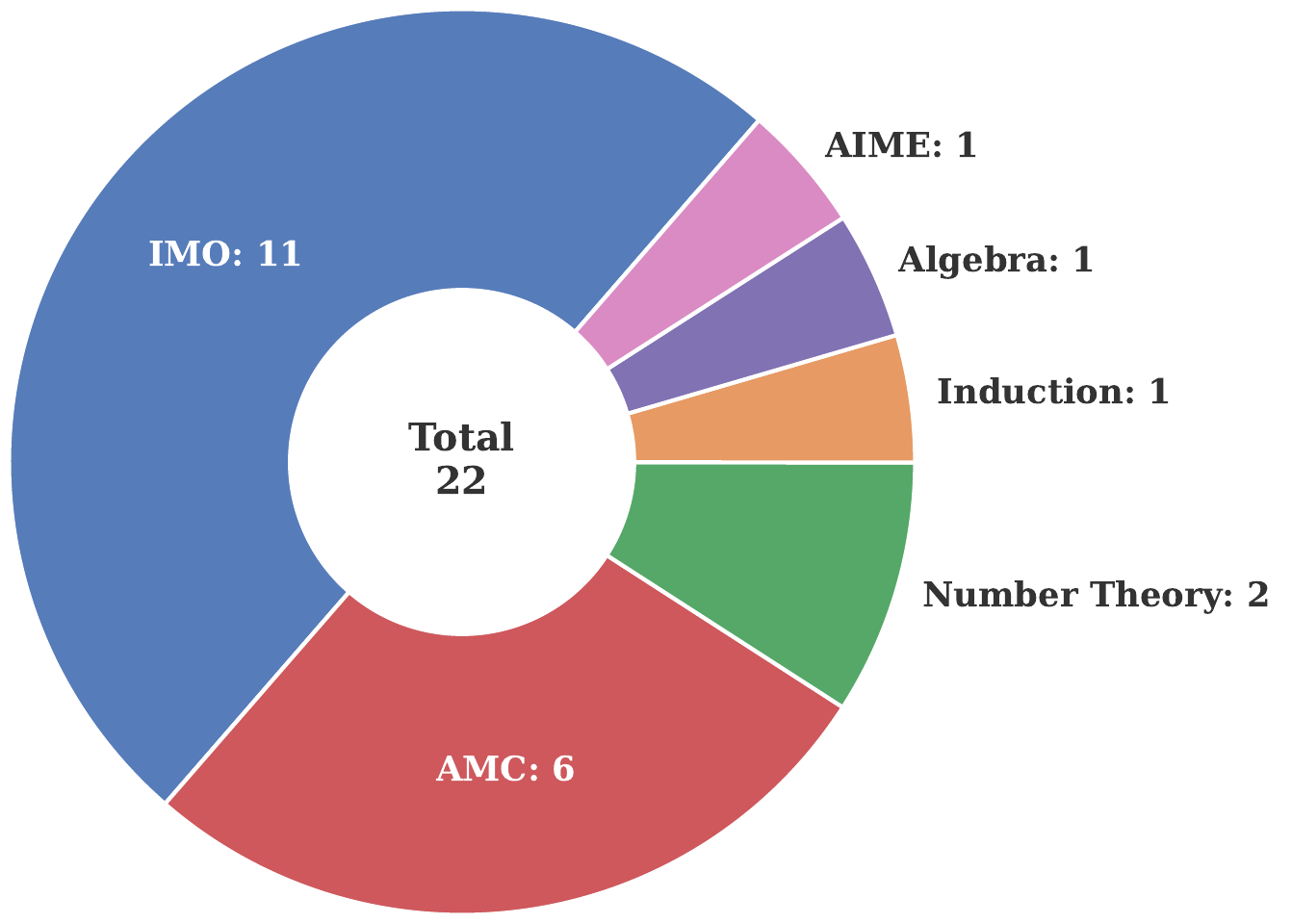}
        \caption{Domain composition of the all-model-wrong set on original MiniF2F-Test.}
        \label{fig:minif2f_ori_piechart}
    \end{subfigure}
    \hspace{0.02\textwidth}
    \begin{subfigure}[t]{0.31\textwidth}
        \centering
        \includegraphics[width=\linewidth,height=0.23\textheight,keepaspectratio]{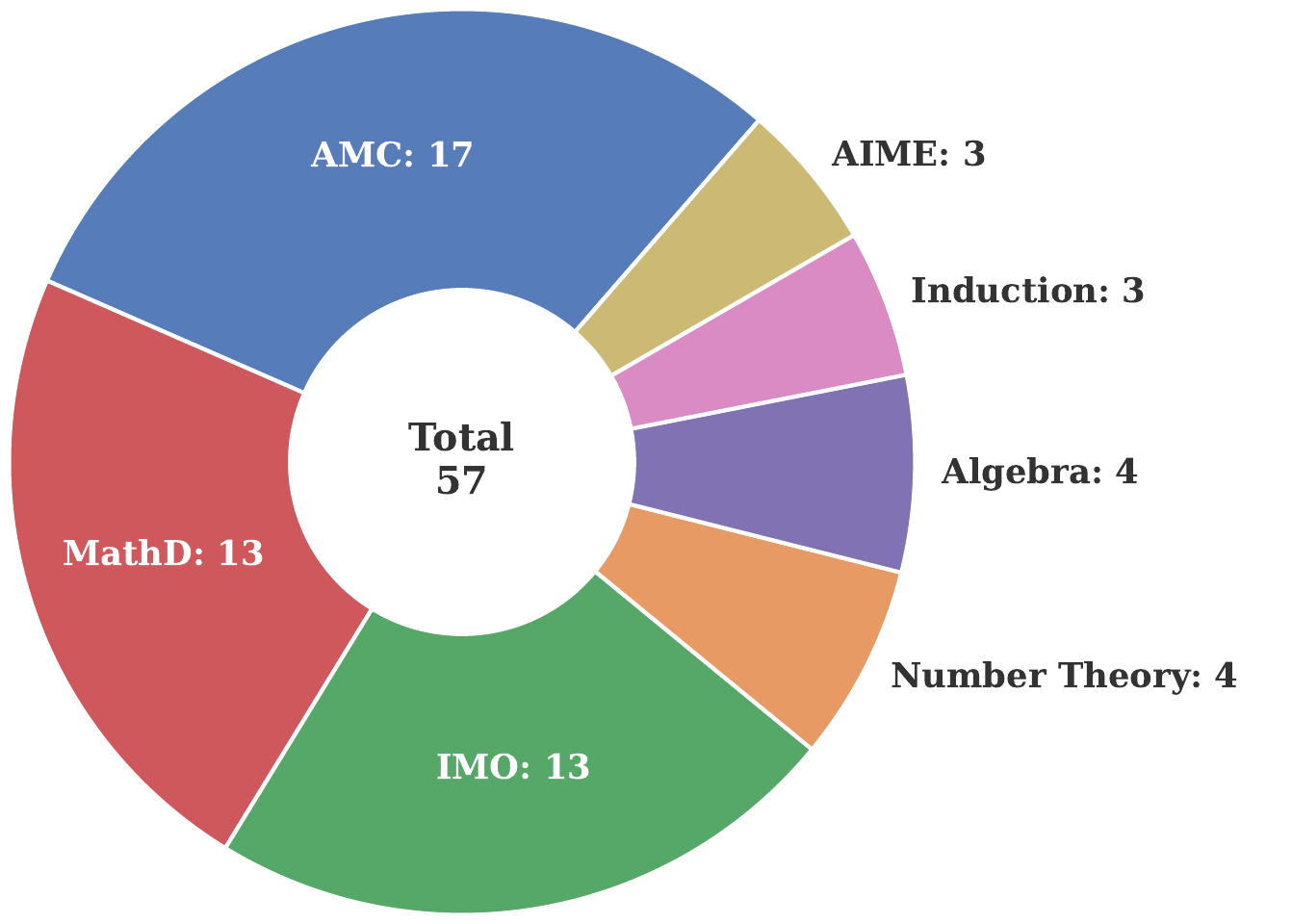}
        \caption{Domain composition of the at-least-one-model-wrong set on MiniF2F-ALF, after applying the ALF mutation scheme to each MiniF2F-Test statement.}
        \label{fig:minif2f_alf_piechart}
    \end{subfigure}
    \caption{
    Categories are taken directly from MiniF2F. The competition-derived slices comprise \emph{IMO} (International Mathematical Olympiad), \emph{AMC} (American Mathematics Competitions, AMC 8/10/12), and \emph{AIME} (American Invitational Mathematics Examination); \emph{MathD} groups problems imported from the MATH~\citep{hendrycks2021measuring}. The remaining slices are topic-tagged hand-written instances grouped by area: \emph{Algebra}, \emph{Number Theory}, and \emph{Induction}.}

\end{figure}

Extending the analysis of \S\ref{sec:minif2f_analysis}, we examine the domain composition of the residual failure set on original MiniF2F-Test under our four-prover panel (\method-4B, \method-32B, Goedel-Prover-V2-8B, Goedel-Prover-V2-32B). Figure~\ref{fig:original_piechart} shows that the at-least-one-model-wrong subset is dominated by olympiad-level material: AMC and IMO together account for roughly two-thirds of all failures, with the remaining categories forming a thin long tail. The all-model-wrong subset (Figure~\ref{fig:minif2f_ori_piechart}) is even more skewed: IMO alone accounts for half of all universal failures and AMC for the bulk of the remainder, so the residual difficulty on which contemporary provers can still be separated sits almost entirely at the upper end of the difficulty spectrum. Combined with the headline saturation statistic that $83.20\%$ of MiniF2F-Test is solved by all four systems, this leaves only a narrow band of olympiad instances on which the benchmark can meaningfully discriminate between strong provers. Figure~\ref{fig:minif2f_analysed} breaks this residual set down per problem, showing for each prover which instances it solves and which it misses, and confirming that disagreement is confined to a smaller, olympiad-heavy tail.

Figure~\ref{fig:minif2f_alf_piechart} shows the same decomposition after ALF mutation. The distribution shifts substantially: MathD, a small slice of the original failure set, rises to roughly $20\%$ of at-least-one-wrong instances on MiniF2F-ALF, and Algebra, Number Theory, Induction, and AIME each occupy a non-trivial share. The IMO/AMC concentration that characterises the original failure set is broken, with failures redistributing across categories that should not be intrinsically harder than instances the same systems solve on the original split. This redistribution is the empirical signature we look for: rather than amplifying the olympiad tail, ALF surfaces a broader set of structurally simple cases on which strong provers turn out to be brittle, providing discriminative signal across the full difficulty range rather than only at its top.

\section{A Failure Mode of Self-Correction}
Self-correction conditions each attempt on the previous Lean error, which repairs mechanical slips but anchors the search around the original proof skeleton: when that skeleton encodes a flawed logical step rather than a typo, every refinement inherits the same flaw. In this section, we provide a representative theorem on which self-correction loops on a logically incorrect lemma chain while independent restart sampling under the same compute budget recovers a valid proof.

\label{app:self-correction-failure}
\begin{tcolorbox}[
  enhanced,
  colback=plotpurple!8, colframe=plotpurple!75!black,
  title={\textbf{Putnam 1966 A4.} Non-square positive integers indexed by closest-square distance.},
  fonttitle=\small,
  left=3pt, right=3pt, top=3pt, bottom=3pt,
]
{\footnotesize\textbf{Formal statement}}
\begin{minted}{lean4}
theorem putnam_1966_a4
    (a : ℕ → ℤ)
    (ha1 : a 1 = 2)
    (hai : ∀ n ≥ 1, a (n + 1) =
        (if ∃ m : ℤ, m^2 = a n + 1 = True
         then a n + 2
         else a n + 1)) :
    ∀ n ≥ 1, a n = n + round (Real.sqrt n) := by
  sorry
\end{minted}
\end{tcolorbox}

\begin{tcolorbox}[
  enhanced,
  colback=plotgreen!8, colframe=plotgreen!75!black,
  title={\textbf{Putnam 1966 A4: \method successful proof.}},
  fonttitle=\small,
  left=3pt, right=3pt, top=3pt, bottom=3pt,
]
\begin{minted}{lean4}
theorem putnam_1966_a4
(a : ℕ → ℤ)
(ha1 : a 1 = 2)
(hai : ∀ n ≥ 1, a (n + 1) =
    (if ∃ m : ℤ, m^2 = a n + 1 = True then a n + 2 else a n + 1))
: ∀ n ≥ 1, a n = n + round (Real.sqrt n) := by
  have h_round_of_block : ∀ (n k : ℕ),
      k * k - k + 1 ≤ n → n ≤ k * k + k →
      round (Real.sqrt (n : ℝ)) = (k : ℤ) := by
    [TRUNCATE]

  have h_square_branch_of_block : ∀ (n k : ℕ),
      k * k - k + 1 ≤ n → n ≤ k * k + k →
      (((∃ m : ℤ, m ^ 2 = (n : ℤ) + (k : ℤ) + 1) = True) ↔ n = k * k + k) := by
    [TRUNCATE]

  have h_block : ∀ n : ℕ, n ≥ 1 → ∃ k : ℕ,
      a n = (n : ℤ) + (k : ℤ) ∧
      k * k - k + 1 ≤ n ∧ n ≤ k * k + k := by
    [TRUNCATE]

  intro n hn
  obtain ⟨k, hak, hl, hu⟩ := h_block n hn
  rw [hak]
  rw [h_round_of_block n k hl hu]
\end{minted}
\end{tcolorbox}

\begin{tcolorbox}[
  enhanced, sidebyside, sidebyside align=top, sidebyside gap=4mm,
  colback=red!4, colframe=red!55!black,
  title={\textbf{Putnam 1966 A4: A self-correction failure mode.} A structural induction on $n$ that decomposes the non-square indices into blocks $[k^2-k+1,\,k^2+k]$, with a rounding lemma identifying $\mathrm{round}(\sqrt{n})$ inside each block.},
  fonttitle=\small,
  left=3pt, right=3pt, top=3pt, bottom=3pt,
]
{\footnotesize\textbf{Goedel-Prover-V2 (first round, failed)}}
\begin{minted}{lean4}
theorem putnam_1966_a4
(a : ℕ → ℤ)
(ha1 : a 1 = 2)
(hai : ∀ n ≥ 1, a (n + 1) = (if ∃ m : ℤ, m^2 = a n + 1 then a n + 2 else a n + 1))
: ∀ n ≥ 1, a n = n + round (Real.sqrt n) :=
by
  have h_main : ∀ (n : ℕ), n ≥ 1 → a n = n + round (Real.sqrt n) := by
    intro n hn
    have h₁ : ∀ n : ℕ, n ≥ 1 → a n = n + round (Real.sqrt n) := by
      intro n hn
      induction' hn with n hn IH
      · norm_num [ha1]
        <;> simp [round_eq, Int.floor_eq_iff, Real.le_sqrt, Real.sqrt_lt]
        <;> norm_num
        <;> linarith [Real.sqrt_nonneg 1,
              Real.sq_sqrt (show 0 ≤ (1 : ℝ) by norm_num)]
      · have h₂ : a (n + 1) =
            (if ∃ m : ℤ, m ^ 2 = a n + 1 then a n + 2 else a n + 1) := by
          apply hai
          <;> omega
        rw [h₂]
        have h₃ : a n = n + round (Real.sqrt n) := IH
        rw [h₃]
        have h₄ : (if ∃ m : ℤ, m ^ 2 = (n : ℤ) + round (Real.sqrt n) + 1
                   then (n : ℤ) + round (Real.sqrt n) + 2
                   else (n : ℤ) + round (Real.sqrt n) + 1)
                = (n + 1 : ℤ) + round (Real.sqrt (n + 1)) := by
          split_ifs with h
          · have h₈ : round (Real.sqrt (n + 1 : ℝ))
                    = round (Real.sqrt (n : ℝ)) + 1 := by
              sorry
            sorry
          · have h₈ : round (Real.sqrt (n + 1 : ℝ))
                    = round (Real.sqrt (n : ℝ)) := by
              sorry
            sorry
        rw [h₄]
    exact h₁ n hn
  intro n hn
  exact h_main n hn
\end{minted}
\tcblower
{\footnotesize\textbf{Self-correction round (failed)}}
\begin{minted}{lean4}
import Mathlib
import Aesop

set_option maxHeartbeats 0

open BigOperators Real Nat Topology Rat Filter

theorem putnam_1966_a4
(a : ℕ → ℤ)
(ha1 : a 1 = 2)
(hai : ∀ n ≥ 1, a (n + 1) = (if ∃ m : ℤ, m^2 = a n + 1 then a n + 2 else a n + 1))
: ∀ n ≥ 1, a n = n + round (Real.sqrt n) := by
  have h_main : ∀ n ≥ 1, a n = (n : ℤ) + round (Real.sqrt n) := by
    intro n hn
    have h₁ : n ≤ 16 ∨ n > 16 := by omega
    cases h₁ with
    | inl h₁ =>
      interval_cases n
      <;> norm_num [ha1, hai, round_eq, Int.floor_eq_iff,
                    Real.sqrt_le_iff, Real.lt_sqrt, Real.sqrt_lt] at *
      <;> (try { nlinarith [Real.sqrt_nonneg 2,
              Real.sq_sqrt (show (0:ℝ) ≤ 2 by norm_num)] })
      <;> (try { nlinarith [Real.sqrt_nonneg 3,
              Real.sq_sqrt (show (0:ℝ) ≤ 3 by norm_num)] })
      <;> (try { nlinarith [Real.sqrt_nonneg 5,
              Real.sq_sqrt (show (0:ℝ) ≤ 5 by norm_num)] })
      <;> (try { aesop }) <;> (try { linarith })
    | inr h₁ =>
      exfalso
      have h₃ : a 1 = 2 := ha1
      have h₄ : a 2 = 3 := by
        have h₅ := hai 1 (by norm_num); norm_num [ha1] at h₅ ⊢
        <;> simp_all [ha1] <;> aesop
      have h₅ : a 3 = 5 := by
        have h₆ := hai 2 (by norm_num); norm_num [h₄] at h₆ ⊢
        <;> simp_all [ha1, h₄] <;> aesop
      have h₆ : a 4 = 6 := by
        have h₇ := hai 3 (by norm_num); norm_num [h₅] at h₇ ⊢
        <;> simp_all [ha1, h₄, h₅] <;> aesop
      have h₇ : a 5 = 7 := by
        have h₈ := hai 4 (by norm_num); norm_num [h₆] at h₈ ⊢
        <;> simp_all [ha1, h₄, h₅, h₆] <;> aesop
      have h₁₈ : a 16 = 20 := by
        have h₁₉ := hai 15 (by norm_num); norm_num [h₁₇] at h₁₉ ⊢
        <;> simp_all [ha1, h₄, h₅, h₆, h₇, h₈, h₉, h₁₀,
                      h₁₁, h₁₂, h₁₃, h₁₄, h₁₅, h₁₆, h₁₇]
        <;> aesop
      <;> norm_num <;> aesop
  intro n hn
  have h₁ : a n = (n : ℤ) + round (Real.sqrt n) := h_main n hn
  norm_cast at h₁ ⊢
  <;> simp_all [add_assoc] <;> linarith
\end{minted}
\end{tcolorbox}

\section{Effective Token Complexity Calculation}
\label{app:etc}

\paragraph{Intuition.}
Wall-clock cost and pass@$N$ are both misleading proxies for compute when two systems decode under very different regimes: pass@$N$ ignores per-attempt length, and wall-clock conflates batching, hardware, and serving stack. We instead estimate the work the attention mechanism performs under an idealised model. For a single decoded block of $a$ tokens conditioned on an input of $m$ tokens, the generated token at within-block position $t$ attends to the $m$ prompt tokens together with the $t$ tokens produced up to and including itself ($m + t$ keys in total), so the cumulative query--key work over the block is

\begin{equation}
\sum_{t=1}^{a} (m + t) \;=\; \underbrace{m\,a}_{\text{prompt attention}} \;+\; \underbrace{\frac{a(a+1)}{2}}_{\text{self-attention}}.
\label{eq:etc}
\end{equation}
We call the quantity in Eq.~\eqref{eq:etc} the \emph{effective token complexity} (ETC),
denoted by $\mathrm{ETC}(m, a)$. The first term scales linearly with the input context (attending to a fixed prompt across the whole block); the quadratic term is the intra-block self-attention cost, and it dominates whenever $a$ is comparable to or larger than $m$. For pass@$N$ we multiply the per-attempt ETC by $N$, summing across rounds for multi-round systems.

\paragraph{Assumptions and scope.}
ETC is deliberately idealised and should be read as a worst-case, asymptotic proxy rather than an estimate of realised inference cost; three caveats apply. First, it assumes \emph{dense global attention} in every layer, so each step pays a cost proportional to its full context length. This does not hold for models that replace global attention with sparse attention (e.g.\ the Qwen3 family) or linear or gated delta net attention (e.g.\ Qwen3.5-Omni) in most layers, for which the per-step context cost is sub-linear and ETC over-counts the linear term. Second, ETC ignores inference-engine optimisations such as KV- and prefix-caching, as well as the memory-bound versus compute-bound trade-off that depends jointly on context length and batch size. Third, it is not a FLOP count: converting it to FLOPs would require architecture-specific factors (model width, head count, attention type), not a single multiplicative constant. What survives these caveats is the \emph{geometric asymmetry} the measure is designed to expose, which holds under any super-linear attention cost: generating many short independent blocks is asymptotically cheaper than conditioning on a long accumulated context. Concretely, $k$ independent blocks of length $n$ cost on the order of $k\cdot n^2/2$ under the self-attention term, against $(kn)^2/2 = k^2 \cdot n^2/2$ for a single block of the same total length, a factor-$k$ saving. Restart sampling realises this saving; multi-round self-correction does not. We therefore use ETC only to characterise this scaling behaviour, and report the per-attempt token statistics in Table~\ref{tab:etc-comparison} so the estimate can be recomputed under any alternative cost model.

\paragraph{Numerical comparison.}
For Goedel-Prover, summing its three self-correction rounds at pass@$184$ with the per-round token statistics in Table~\ref{tab:etc-comparison}, we obtain

\[
\begin{aligned}
\mathrm{ETC}_{\text{Goedel-Prover}}
&= 184\,\Big[\;
   \underbrace{\mathrm{ETC}(284.88,\,18367.83)}_{\textit{round 1}} %
 + \underbrace{\mathrm{ETC}(7838.74,\,21638.70)}_{\textit{round 2}} %
 + \underbrace{\mathrm{ETC}(8309.06,\,23835.80)}_{\textit{round 3}} \;\Big] \\[0.6em]
&\approx 184\,\Big[\,
   1.7393{\times}10^{8}
 + 4.0375{\times}10^{8}
 + 4.8214{\times}10^{8} \Big] \\
&\approx 1.95\times 10^{11}.
\end{aligned}
\]
For \method, with an input context of $284.88$ tokens and an average restart length of $18{,}741.80$ tokens at pass@$1024$,
\[
\begin{aligned}
\mathrm{ETC}_{\text{\method}}
&= 1024 \cdot \mathrm{ETC}(284.88,\,18741.80) \\
&\approx 1024 \cdot 1.8098{\times}10^{8}
\;\approx\; 1.8532\times 10^{11}.
\end{aligned}
\]

With these inputs, $\mathrm{ETC}_{\text{\method}} / \mathrm{ETC}_{\text{Goedel-Prover}} \approx 0.95$, so \method incurs roughly $5\%$ lower effective token complexity than Goedel-Prover under this approximation. Although \method runs far more attempts ($1024$ vs.\ $184$), each restart conditions on a short input context, whereas Goedel-Prover's later self-correction rounds operate over much longer accumulated contexts. The quadratic-in-$a$ term, combined with the larger $m$ in those later rounds, outweighs \method's higher attempt count.

\end{document}